# AI for Computational Design Science: A Responsible Human-AI Framework and Case Study on Short-Form Video Safety Surveillance

Wenli Zhang
Iowa State University
3332 Gerdin, 2167 Union Drive
Ames, IA, USA 50011
Email: wlzhang@iastate.edu

Jiaheng Xie
University of Delaware
217 Purnell Hall
Newark, DE, USA 19716
Email: jxie@udel.edu

Zhihe Pan
City University of Hong Kong
83 Tat Chee Avenue
Kowloon Tong, Kowloon, Hong Kong
Email: zhihepan2-c@my.cityu.edu.hk

Yidong Chai
City University of Hong Kong
83 Tat Chee Avenue
Kowloon Tong, Kowloon, Hong Kong
Email: yidong.chai@cityu.edu.hk

Xiao Fang
University of Delaware
217 Purnell Hall
Newark, DE, USA 19716
Email: xfang@udel.edu

Sudha Ram
University of Arizona
McClelland Hall 430J, 1130 E. Helen St.
Tucson, Arizona 85721
Email: sram@arizona.edu

Please send comments to Wenli Zhang at wlzhang@iastate.edu.

# AI for Computational Design Science: A Responsible Human-AI Framework and Case Study on Short-Form Video Safety Surveillance

**Abstract**: Artificial Intelligence (AI) is beginning to transform not only what information systems researchers design, but also how they conduct design research. Yet existing literature offers limited guidance on how AI actively participates in the key stages of computational design science (CDS) research, including problem formulation, resource construction, design search, evaluation, and knowledge abstraction. This study develops AI for Computational Design Science (AI4CDS), a five-phase methodological framework for conducting CDS when AI becomes a constitutive participant in the research process. AI4CDS specifies how AI expands the search for problems and designs, while researchers retain responsibility for domain grounding, admissibility, verification, and scientific judgment. Their collaboration is governed by graduated trust, reversibility, and auditability, with differentiated reproducibility providing a practical framework for conducting verification and maintaining auditability. AI4CDS is instantiated and examined through the development of ChildRiskGuard, an interpretable computational artifact for detecting short-form videos that are inappropriate for children; the corresponding AI interactions, rejected alternatives, corrections, and audit trails are documented in Online Appendix A. Following AI4CDS, this case study translates audience-dependent safety and explanation faithfulness into three technical challenges and, through AI-expanded and researcher-governed design search, develops an artifact that separates generic from child-specific risk, represents distinct developmental-risk mechanisms, and makes concept-level explanations part of the predictive computation. ChildRiskGuard achieves an F1 score of 0.769, substantially outperforming the direct application of a general-purpose content-safety model to child-appropriateness detection while remaining competitive with strong alternative benchmarks. The primary contribution of this work is AI4CDS as a responsible methodological framework for AI-enabled CDS; ChildRiskGuard provides consequential process and artifact evidence showing how the framework can generate, evaluate, and abstract novel computational design knowledge.

## 1. Introduction

Recent Information Systems (IS) scholarship recognizes that Artificial Intelligence (AI) may reshape not only research outputs but also the conduct of scientific inquiry itself, including design-oriented research (Gopal et al. 2025). For computational design science (CDS), which addresses consequential business and societal problems through novel computational models, algorithms, and methods (Rai 2017, Fang et al. 2025), this raises a fundamental methodological question: how should CDS be conducted when AI becomes a constitutive participant in producing artifacts and design knowledge? Existing CDS guidance explains how researchers motivate domain-specific problems, construct and evaluate computational artifacts, and abstract design knowledge, but does not specify how AI participation should be organized across the CDS lifecycle (Hevner et al. 2004, Gregor and Hevner 2013, Padmanabhan et al. 2022, Abbasi et al. 2024, Fang et al. 2025). Human–AI research frames interaction with increasingly agentic AI as a delegation problem and shows that effective collaboration depends on responsibility allocation and capability assessment (Baird and Maruping 2021, Fügener et al. 2022), but does not explain how such delegation should be structured across the interdependent activities of CDS. AI can expand problem formulations, explore methodological alternatives, construct computational resources, support evaluation, and broaden design search, while human researchers remain responsible for domain grounding, sociotechnical and ethical judgment, verification, and scientific accountability. AI also introduces risks of hallucination, stochasticity, information leakage, and weakened provenance. The resulting question is therefore not simply how to design AI artifacts, but how human and AI capabilities, authority, and accountability should be organized across CDS.

We address this gap and propose **AI for Computational Design Science (AI4CDS)**, a five-phase responsible human–AI framework that specifies how human and AI roles co-evolve across the CDS lifecycle (Figure 1). In Phase 1, AI-Expanded Research Problem Formulation: AI maps and synthesizes relevant knowledge, while human researchers ground and scope the consequential problem and identify domain-specific technical challenges. In Phase 2, AI-Augmented Data & Resource Construction: AI explores candidate data, representations, and computational resources, while humans validate their domain meaning and admissibility. In Phase 3, AI-Generative Research Design Search & Refinement: AI generates and critiques alternative designs, while humans define the admissible design space and commit the scientific design decisions. In Phase 4, AI-Enabled Multi-Faceted Evaluation: AI supports comparative analysis and diagnosis, while humans fix the evaluation protocol, verify the evidence, and judge artifact utility and design claims. In Phase 5, AI-Assisted Design Principle Distillation: AI organizes provenance and surfaces cross-phase patterns, while humans determine which findings warrant abstraction into transferable design knowledge. Across the research process, differentiated reproducibility matches different research activities to appropriate verification standards. Across all five phases, AI expands the breadth of scientific search, whereas humans govern scientific commitments. Graduated trust, reversibility, and auditability constrain this collaboration throughout the lifecycle.

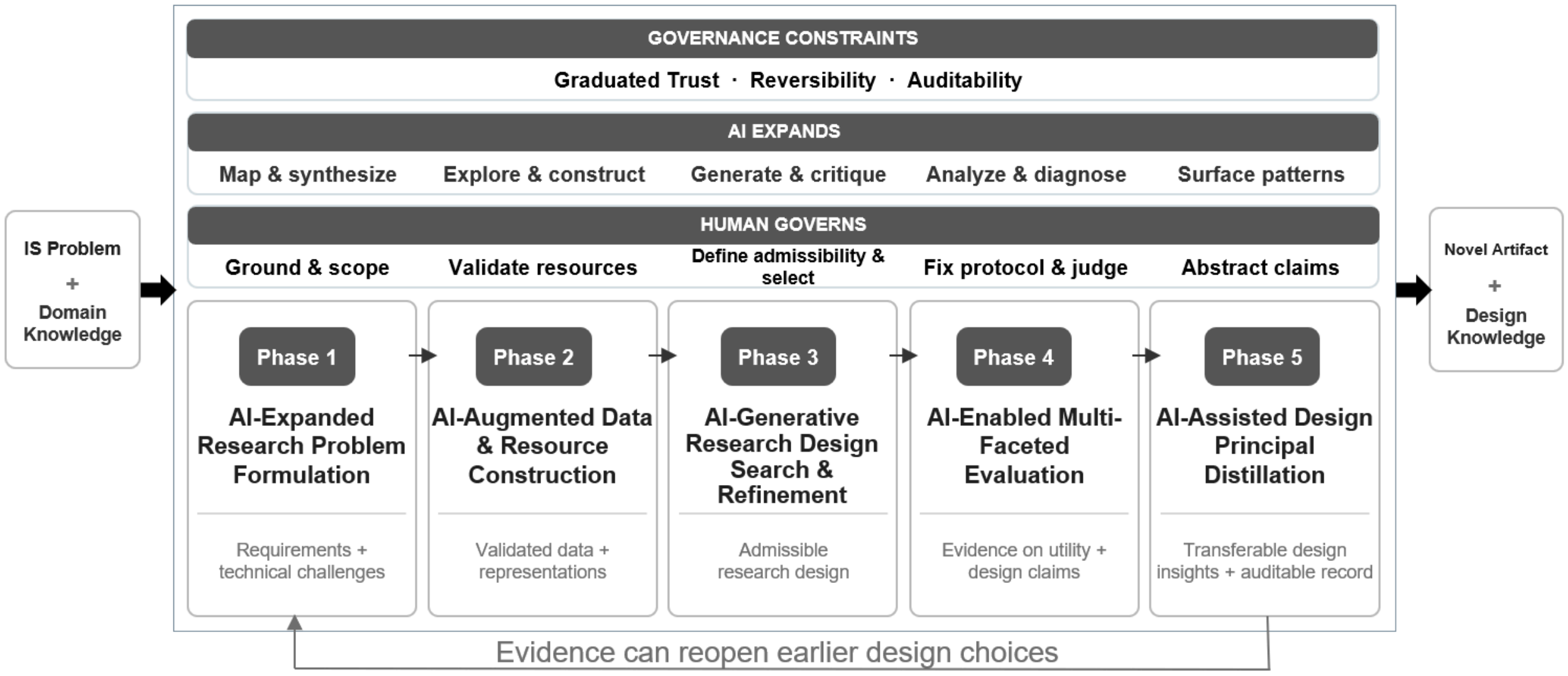

**Figure 1. AI4CDS: A Responsible Human–AI Framework for Computational Design Science**

A meaningful examination of AI4CDS requires a consequential design problem in which AI cannot simply optimize a well-specified task, because domain knowledge plays a central role in determining what constitutes an acceptable computational solution. Child-oriented short-video safety provides such a setting. Content that is acceptable for a general audience may still be developmentally inappropriate for children, and seemingly benign visual conventions may conceal harmful material (Balanzategui 2019, Papadamou et al. 2020). Therefore, general-purpose harmful-content detection cannot be directly transferred to this context. Instead, the research process must translate domain knowledge about children, safety, and explanation into computational requirements that guide and constrain the design space.

To instantiate AI4CDS in this setting, we develop ChildRiskGuard: an interpretable safety surveillance system for identifying child-inappropriate content in short-form videos. In Phase 1, we identify two higher-order gaps. First, safety is typically treated as a property of content rather than in relation to the audience. Second, explanations are often produced separately from the predictions they are intended to explain. These gaps are translated into three technical challenges: preserving generic safety capability while isolating additional child-specific risk, representing heterogeneous developmental risk mechanisms, and ensuring concept-level faithfulness. Subsequent AI4CDS phases use these requirements to construct resources, expand and discipline the design search, evaluate alternatives, and ultimately develop ChildRiskGuard. ChildRiskGuard thus serves as the consequential artifact through which AI4CDS is instantiated and examined, with its development process providing evidence of how the framework structures domain-informed computational design. Table 1 traces how each AI4CDS phase is instantiated in the ChildRiskGuard study, linking the framework's human–AI division of labor to the concrete research decisions and outputs that constitute the case evidence.

**Table 1. From AI4CDS Framework to ChildRiskGuard Instantiation**

| AI4CDS Phase | Existing CDS Emphasis | → | AI4CDS Added Prescription (Human–AI Division of Labor) | ChildRiskGuard Instantiation and Evidence |
|---|---|---|---|---|
| Phase 1: AI-Expanded Research Problem Formulation | Formulate a consequential problem | → | ● AI maps and synthesizes problem and solution knowledge.<br>● Researchers verify evidence, scope child-oriented safety, and derive domain requirements and technical challenges. | Audience-dependent safety and faithful explanation identified as the two higher-order gaps, yielding three technical challenges: isolating child-specific risk, representing heterogeneous developmental mechanisms, and preserving concept-level faithfulness. |
| Phase 2: AI-Augmented Data & Resource Construction | Construct data and computational resources | → | ● AI helps explore candidate representations and measurement resources.<br>● Researchers determine their domain validity and fix the resources admitted into the artifact. | Frozen LlavaGuard generic-safety capability and evidence channels; three knowledge-grounded developmental-risk mechanisms; deterministic and vision-language model (VLM)-derived measurements and normalization procedures. |
| Phase 3: AI-Generative Research Design Search & Refinement | Search and construct computational designs | → | ● AI generates and critiques alternative architectures, objectives, and constraints.<br>● Researchers define admissibility and select theoretically justified, falsifiable designs. | Four coordinated design elements: identifiable child-specific residual learning, knowledge-anchored mechanism structuring and gating, a semantic contribution budget, and axiomatically faithful monotone additive hazard mapping. |
| Phase 4: AI-Enabled | Evaluate artifact utility | → | ● AI supports implementation, benchmarking, ablation, | Predictive evaluation on 7,070 videos; comparisons across four benchmark |

| Multi-Faceted Evaluation | and design claims | | diagnostics, and robustness analysis.<br>● Researchers lock the protocol, verify implementation, and interpret the evidence. | families; component- and mechanism-level ablations; explanation-faithfulness cases; robustness and generalizability analyses. |
|---|---|---|---|---|
| Phase 5: AI-Assisted Design Principle Distillation | Abstract and communicate design knowledge | → | ● AI organizes provenance and cross-phase evidence and surfaces recurring patterns.<br>● Researchers determine what can be generalized beyond the focal case. | Auditable AI-use record; transferable design insights; graduated trust, reversibility, and auditability principles; differentiated reproducibility guidance and boundary conditions. |

The primary contribution of this work is AI4CDS, a framework for CDS that organizes and governs constitutive AI participation across the research lifecycle. AI4CDS extends CDS by specifying how AI expands scientific search across problem formulation, resource construction, design search, evaluation, and knowledge abstraction, while researchers retain authority over domain grounding, admissibility, verification, and scientific commitments and accountability. ChildRiskGuard provides the consequential artifact and process evidence through which AI4CDS is instantiated and examined. Its development demonstrates how the framework can translate domain knowledge into computational requirements and use these requirements to guide the design of an interpretable AI system. Substantively, the study also advances child-oriented short-video safety by showing how developmental considerations can be incorporated into harmful-content detection and explanation. More broadly, AI4CDS provides a foundation for future IS researchers seeking to use AI not merely alongside CDS, but as a responsible participant in it.

## 2. Related Work

This section implements Phase 1 of AI4CDS: AI-Expanded Research Problem Formulation. This phase begins with researcher-led, AI-supported (1) problem scoping and (2) jointly reviews descriptive knowledge on child risks and prescriptive knowledge on inappropriate-video detection, interpretable prediction, and concept bottlenecks. (3) AI broadens literature retrieval and organization, while (4) researchers verify sources, synthesize the evidence, and assess what existing approaches can and cannot address. The verified evidence is then used to (5) define the consequential problem and (6) identify domain requirements and translate them into technical challenges for design. In ChildRiskGuard, this process reveals the need to preserve generic safety while identifying additional child-specific risk, represent heterogeneous developmental-risk mechanisms, and maintain faithful concept-level explanations. (7) When emerging requirements expose missing knowledge, the literature review is reopened. Detailed procedures and audit records are articulated in Online Appendix A.1.

### 2.1. Audience-Specific Video Safety and Inappropriate-Video Detection

Most content-moderation systems operationalize harm primarily from characteristics of the content, whereas appropriateness for children is inherently audience-dependent. Content that is acceptable for a general audience may nevertheless be inappropriate for children because the consequences of exposure depend on developmental characteristics of the viewer (Papadamou et al. 2020). Developmental research provides a substantive basis for this distinction: young children differ from adults in how they transfer and interpret mediated information and respond to screen content (Barr 2010, Guellai et al. 2022).

This distinction creates an important computational requirement. Child-specific risk may arise even when a video contains none of the explicit signals targeted by general safety systems, and such content may deliberately adopt visual conventions associated with children's media (Balanzategui 2019, Papadamou et al. 2020). Existing moderation systems therefore provide useful information about generic safety but cannot serve as proxies for child appropriateness (Ahmed et al. 2023). A child-oriented detector must preserve useful generic-safety information while identifying *additional risk that arises specifically because the audience is children*. This requirement cannot be satisfied simply by lowering the decision threshold of a general-purpose safety detector.

Existing inappropriate-video detection methods span reconstruction-based deviation scoring (Hasan et al. 2016), bag-level segment ranking (Sultani et al. 2018), cross-modal fusion, including child-oriented moderation (Ahmed et al. 2024), and vision-language scoring or judgment (Tang et al. 2024). These approaches share two limitations for child-oriented moderation. First, they generally define risk from content without explicitly representing the audience for whom the content is harmful. Consequently, they do not distinguish generic safety risk from risk that arises specifically for children. Second, existing approaches provide limited support for explanations that are demonstrably tied to the computation producing the moderation decision. Vision-language models (VLMs) can generate natural-language rationales, but because rationale generation is separate from the prediction itself, those rationales may not reflect the evidence that actually drove the decision.

For our setting, these limitations shift the design question from choosing among existing detector families to identifying which capabilities can serve as building blocks for child-oriented moderation. Reconstruction and bag-ranking methods are poorly matched to supervised short-video moderation, while cross-modal fusion incorporates relevant modalities but does not naturally expose the domain-specific source of a decision. VLMs are more suitable as a measurement foundation because pretrained representations provide rich semantic information and a frozen safety model can provide a generic-safety reference. We therefore reuse frozen vision-language capabilities for representation and generic-safety measurement rather than as an end-to-end child-appropriateness classifier.

This choice leaves the explanation problem unresolved. For child-oriented moderation, we require the prediction to be structured around named developmental-risk concepts such that the reported explanation is derived from the same quantities that produce the decision, rather than generated after the fact. We therefore next examine interpretable methods for video prediction.

### 2.2. Interpretable Video Prediction

Interpretable audio-visual video classification methods place explanations at different points relative to prediction, including forward-pass attention weights (Nagrani et al. 2021), post-hoc saliency maps (Selvaraju et al. 2017), localized visual or audio grounding (Arandjelović and Zisserman 2018), concept- or additive-decomposition (Kumar et al. 2025), and natural-language justifications (Park et al. 2018). The key distinction among these approaches is not whether they provide an explanation, but whether the explanation reflects the information that the model actually used to make its prediction. Attention and gradient-based methods expose aspects of model behavior, but the exposed weights or saliency patterns need not constitute the evidence sufficient for the prediction (Adebayo et al. 2018, Jain and Wallace 2019). Grounding methods more directly connect representations to localized evidence, but localization of an object or sound does not necessarily explain why that evidence produced a particular downstream risk judgment. Natural-language methods can express domain-relevant reasons in an accessible form, but generated rationales may remain computationally decoupled from the prediction they describe. Concept-based methods offer a different possibility: named concepts can be placed directly on the predictive pathway so that prediction and explanation operate over the same intermediate variables.

Two requirements are especially important for child-oriented moderation. First, the explanation should identify the specific developmental-risk mechanism underlying the prediction, rather than only highlighting pixels, frames, or modalities. For example, a video involving imitable dangerous actions and one involving self-harm may require different forms of review or intervention. Second, simply including human-interpretable concepts with predefined meanings (such as skin exposure, risky actions, or self-harm evidence) in the model does not guarantee faithful explanations. A concept may still be influenced by information unrelated to its intended meaning or may capture additional information about the target label (Margeloiu et al. 2021, Havasi et al. 2022). Thus, an interpretable architecture must control both what information enters through a named concept and how that concept contributes to the final prediction.

These requirements make concept-based prediction the most appropriate foundation for our study. Unlike attention or saliency, concepts can represent developmental mechanisms directly; unlike post-hoc natural-language rationales, they can participate in the computation that produces the decision. We therefore adopt a concept-based architecture in which vision-language representations provide semantic measurements while named developmental concepts structure the child-specific risk assessment. Concept bottleneck models provide the most relevant methodological foundation for this design.

### 2.3. Concept Bottleneck Models

Concept bottleneck models (CBMs) place human-understandable concepts between the input representation and the final prediction, making the concept layer part of the predictive pathway (Koh et al. 2020). Subsequent variants enrich how concepts are represented, including embedding-based and probabilistic formulations (Zarlenga et al. 2022, Kim et al. 2023). This structure is attractive for our setting because variables used to explain a moderation decision can participate directly in producing it.

Existing CBMs obtain concepts primarily from human annotation or automated concept generation. Human-annotated CBMs specify concepts in advance but require sample-level concept labels, while more recent methods use pretrained language and VLMs to propose or measure concepts (Oikarinen et al. 2022, Srivastava et al. 2024). Automated generation improves scalability, but neither approach by itself establishes that the concept vocabulary represents the mechanisms through which content becomes harmful to a particular audience. We therefore ground the concept vocabulary in external domain knowledge rather than deriving its substantive meaning from the training samples. As shown in Table 2, prior developmental research identifies mechanisms through which mediated content may create distinctive risks for children, including observational imitation, sexualization, and self-harm contagion (Bandura et al. 1961, Phillips 1974, APA 2007). This substantive knowledge defines the mechanisms represented in the primary model and determines the conceptual organization of the artifact independently of model fitting. Pretrained models and deterministic extractors are then used to measure observable evidence associated with these theoretically defined mechanisms. In this way, domain knowledge determines what the artifact should represent, while computational models determine how evidence for each construct is measured. Table 2 presents the three theory-grounded mechanisms used in the ChildRiskGuard instantiation (without claiming a comprehensive taxonomy of child-inappropriate content.) The architecture supports an extensible set of externally grounded mechanisms, as demonstrated in Online Appendix C.1 with the addition of a fourth mechanism (M4).

**Table 2. Theory-Grounded Mechanisms in the Primary ChildRiskGuard Instantiation**

| Theoretical Basis | Child-specific Risk | Design Implication | Mechanism | Operationalization in ChildRiskGuard |
|---|---|---|---|---|
| Observational learning (Bandura et al. 1961) | Children may imitate dangerous actions observed in the media. | Identify imitable actions using motion, pose, and semantic evidence. | M1: Imitable dangerous actions | Primarily identified using the motion family: body motion, global flow, flow peak, and motion energy.[a] |
| Sexualization research (APA 2007) | Adultized or sexualized performances may be inappropriate for children. | Combine physical exposure with adultized performance context. | M2: Sexualized or adult-like performance | Primarily identified using two evidence families: skin exposure and five adultized-performance prompts.[a] |
| Self-harm contagion (Phillips 1974) | Self-harm content may produce imitation or contagion effects. | Identify self-harm and related emotional or visual cues. | M3: Self-harm or depression-related content | Primarily identified using the low-brightness family, which captures visual darkness.[a,b] |

**Note**: [a] Additional concepts contribute to mechanism intensity but are not sufficient on their own to establish the presence of the corresponding mechanism. These include pose inversion for M1, the broad suggestive-recall concept for M2, and self-harm or depression-related semantic concepts for M3. [b] For M3, self-harm and depression-related semantic concepts serve as intensity-modulating evidence, while low brightness provides the primary evidence used to determine mechanism presence.

A natural alternative is to train a child-specific classifier directly from the overall child-appropriateness label. Although such a model can predict whether a video is appropriate for children, it does not reveal the developmental-risk mechanisms underlying the prediction. Multitask learning offers a more explicit alternative by jointly predicting the overall outcome and specific risks, such as observational imitation, sexualization, and self-harm, through a shared representation (Caruana 1997). However, when these mechanism-level risks are learned in parallel with the final outcome, they mainly serve as auxiliary predictions and do not necessarily determine how the final moderation decision is made. A stronger connection can be established by explicitly deriving the final child-appropriateness prediction from the predicted mechanisms or concepts, as in a CBM (Koh et al. 2020). This formulation raises a central question: *whether developmental-risk mechanisms can form an identifiable and faithful pathway from video content to the final moderation outcome*. Implementing such a pathway through conventional multitask or concept-based learning typically requires sample-level annotations for each developmental-risk mechanism. Such mechanism-level annotations are costly to obtain, and the labeling burden increases substantially as the number of developmental-risk mechanisms expands. ChildRiskGuard addresses this limitation by using externally grounded developmental-risk mechanisms and concept measurements to structure the predictive pathway while learning only from the overall child-appropriateness label. As a result, the model can quantify mechanism-level and concept-level contributions as part of the same computation that produces the final prediction, without requiring mechanism-level supervision.

Standard CBMs nevertheless leave three unresolved requirements in our setting. First, they generally construct a prediction from the concept layer rather than formalizing incremental risk relative to an existing predictor. Child-oriented moderation therefore requires a model that preserves generic-safety information while separately representing the developmental-risk mechanisms that contribute additional risk for children. Second, CBMs commonly organize concepts within a single predictive layer, whereas child-inappropriateness can arise through substantively different developmental mechanisms that require separate representation and attribution. Third, a concept bottleneck does not by itself guarantee faithful concept-level explanations. Predicted concept values may encode information beyond their declared semantics (Margeloiu et al. 2021, Havasi et al. 2022), while highly expressive concept-to-label mappings can make individual concept contributions difficult to identify. The artifact must therefore address both input-side semantic leakage and mapping-side attribution. These three limitations correspond directly to the technical challenges we developed in the next subsection.

### 2.4. Research Gaps and Design Challenges

Our literature review reveals two higher-order gaps. First, inappropriate-video detection methods generally treat harm as a property of the content and do not explicitly represent the audience for whom the content is harmful. This formulation is inadequate for child-oriented moderation because content that is acceptable for a general audience may nevertheless pose developmentally specific risks to children. Second, existing interpretable video models do not necessarily ensure that human-interpretable concepts with predefined meanings capture only the evidence they are intended to represent, or that their reported contributions accurately reflect how those concepts influence the final prediction. In child-safety moderation, this separation is consequential because moderators may take different actions depending on the developmental risk attributed to a video. An appropriate artifact must therefore distinguish generic safety risk from additional child-specific risk and represent the latter through interpretable developmental mechanisms that directly participate in prediction. These gaps identify concept-based prediction as the appropriate methodological foundation while showing why a standard CBM is insufficient. They give rise to three technical challenges that structure the artifact design in the research design section.

**Technical Challenge 1: Extending Generic Safety with Identifiable Child-Specific Risk**. A child-specific model could be trained directly using child-appropriateness labels. However, this approach would not distinguish between risks that are already captured by a general-purpose safety model and risks that arise specifically because the intended audience is children. We therefore use generic safety as an explicit reference channel and design the child-specific pathway to capture only the additional risk associated with child viewers. This design must address three challenges. First, generic and child-specific risks should be combined without allowing one to offset the other. Second, the generic safety signal may need to be calibrated for the child-oriented moderation context. Third, the child-specific component should avoid reproducing information already represented by the generic safety channel. The resulting model should therefore preserve useful general safety information while learning an identifiable child-specific component that represents incremental, rather than redundant, risk.

**Technical Challenge 2: Representing Heterogeneous Child-Specific Risk Mechanisms.** Table 2 identifies three salient, knowledge-grounded mechanisms used in our primary model: imitable dangerous actions, sexualized or adult-like performance, and self-harm or depression-related content. These mechanisms reflect different developmental concerns, rely on different types of evidence, and may occur at substantially different frequencies. Treating them as a single undifferentiated set of concepts can therefore be problematic. In a flat CBM, all concepts are processed through the same prediction function, which may obscure less frequent but consequential mechanisms and provides limited support for mechanism-specific attribution and evaluation. Moreover, the mechanism structure should ideally be learnable from overall child-appropriateness labels, because obtaining sample-level annotations for the specific developmental-risk mechanism associated with each video is typically costly. To address this limitation, the artifact requires a modular structure that separates evidence associated with distinct developmental mechanisms and preserves each mechanism's contribution to the moderation decision. The architecture allows additional externally grounded mechanisms to be incorporated.

**Technical Challenge 3: Preserving Faithfulness across the Concept Bottleneck.** Faithfulness can fail at two distinct points in a concept-based architecture. On the *input side*, semantic concept scores obtained from pretrained representations may encode label-relevant information beyond their declared meanings, creating concept leakage. This risk is particularly relevant when concepts are measured using frozen vision-language representations rather than exhaustive concept-level annotations (Oikarinen et al. 2022, Srivastava et al. 2024). The artifact must therefore retain useful semantic context while limiting the extent to which potentially leaky semantic measurements can determine the prediction.

On the *mapping side*, exposing concept values does not guarantee that their reported contributions faithfully represent the resulting decision. A simple linear mapping facilitates attribution but may be too restrictive for risk relationships involving thresholds or saturation, whereas an unconstrained nonlinear mapping can improve flexibility at the expense of identifiable contributions. We therefore define four faithfulness axioms for the concept-to-risk mapping: A1 (completeness), requiring concept contributions to reconstruct the corresponding mechanism risk; A2 (sign consistency), requiring contributions to follow their theoretically specified direction; A3 (unique attribution), requiring each contribution to be assigned to a specific concept and mechanism; and A4 (non-masking), requiring positive child-specific risk not to be canceled elsewhere in the pathway. Proposition 4 later establishes that the proposed architecture satisfies these four axioms.

## 3. Research Design

This section implements Phase 2: AI-Augmented Data & Resource Construction and Phase 3: AI-Generative Research Design Search & Refinement. In Phase 2, AI explores resources, while researchers (1) test construct validity, (2) inspect measurement behavior, (3) assign signals to general safety, contextual attenuation, mechanism activation, or intensity, and (4) freeze normalization and thresholds using training data. These checks yield ChildRiskGuard's frozen LlavaGuard safety reading and mechanism-aligned concept measurements. Phase 3 then (1) translates the three technical challenges into testable requirements, (2) uses AI to generate component alternatives, (3) pairs candidates with ablations, diagnostics, or theoretical tests before selection, (4) critically reviews claims and constructions, and (5) records consequential decisions. Researchers retain, revise, or reject alternatives, yielding four elements: identifiable residual learning, mechanism gating, semantic budgeting, and faithful additive hazard mapping. Detailed human–AI interactions, candidate dispositions, and revision records are documented in Online Appendix A.2–A.3.

### 3.1. Computational Problem Formulation

We formulate child-oriented short-video moderation as a computational design problem that requires both identifying videos inappropriate for children and explaining the underlying risks. Child appropriateness overlaps with, but is distinct from, generic content safety: some risks are shared, while others arise from child-specific developmental mechanisms and may not be captured by generic safety categories. Moreover, generic safety knowledge is not necessary for predicting child appropriateness, as a sufficiently flexible model could learn directly from child-specific labels. Our design objective is therefore to retain useful generic safety capabilities while explicitly and interpretably learning additional child-specific risks. This decomposition allows the artifact to distinguish generic safety concerns from child-specific risks and identify the concepts contributing to the child-specific assessment.

Let $D = \{(V_i, y_i)\}_{i=1}^{D}$ denote a dataset of $D$ short-form videos, where $y_i \in \{0, 1\}$ indicates whether video $V_i$ is inappropriate for children. Each video carries a visual track and an audio track, and we write $V_i = (V_i^{vis}, V_i^{aud})$. Because risk evidence may occur only during a limited portion of a video, the visual track is represented by $T$ uniformly sampled frames, $X_i = \{x_{i,1}, ..., x_{i,T}\}$.

A fixed measurement function $M$ transforms the video into a generic-safety reading $u_i \in R_+$, interpretable concept measurements $c_i \in R^K$, and frozen generic-safety evidence channels $s_i \in R^L$:

$$(u_i, c_i, s_i) = M(V_i). \tag{1}$$

Given these fixed measurements, a trainable risk-assessment function $H_\theta$ produces a nonnegative generic-safety hazard $\lambda_i^G$, $R$ nonnegative child-specific mechanism hazards $\lambda_i^r$, and an input-independent nonnegative baseline hazard $\lambda^0$. The total hazard is

$$\Lambda_i = \lambda_i^G + \sum_{r=1}^{R} \lambda_i^r + \lambda^0. \tag{2}$$

The total hazard is mapped monotonically to a child-inappropriateness probability $p_i = \rho(\Lambda_i) \in [0, 1]$, and the moderation decision is $\hat{y}_i = I(p_i \geq \eta)$, where $\eta$ is an operating threshold.

This formulation separates the artifact into two stages. The *measurement stage* $M$ determines what generic-safety and concept evidence is present in the video. The *risk-assessment stage* $H_\theta$ determines how those measurements should be transformed, organized, and combined into child-oriented risk. Only $H_\theta$ is learned from the child-inappropriateness labels. Keeping $M$ fixed ensures that the measurement mapping does not change during training. The concept measurements $c_i$ and the generic-safety reading $u_i$ enter the prediction, whereas $s_i$ enters the training objective only, where it serves as the reference against which child-specific components are kept separable.

Let $u$, $C$, $S$, and $y$ collect the generic-safety readings, concept measurements, generic-safety evidence channels, and observed labels in the training data. The learning problem is

$$\theta^* = arg\,min_\theta L(\theta; u, C, S, y). \tag{3}$$

The formulation imposes one baseline performance requirement and three design requirements. Predictive effectiveness requires accurate discrimination between child-inappropriate and child-appropriate videos, while the three technical challenges in Section 2.4 define the admissibility criteria for subsequent design search. Under Phase 3 of AI4CDS, candidate designs were evaluated against these criteria and retained, revised, or rejected accordingly. The four design elements described next are the resulting artifact-level outcomes of that process.

### 3.2. ChildRiskGuard

ChildRiskGuard is a mechanism-aligned concept-bottleneck additive-hazard model comprising a frozen measurement stage and a trainable child-oriented risk-assessment stage. Computationally, the artifact follows

$$X_i \xrightarrow{M} (u_i, \tilde{c}_i),\ (u_i, \tilde{c}_i) \xrightarrow{H_\theta} \{\lambda_i^G, \lambda_i^1, ..., \lambda_i^R, \lambda^0\} \rightarrow \Lambda_i \rightarrow p_i. \tag{4}$$

The AI4CDS design search ultimately retained four coordinated design elements: separating generic from child-specific risk, structuring heterogeneous developmental-risk mechanisms, constraining potentially leaky VLM-derived semantic evidence, and integrating concept-level contributions directly into prediction. Each element corresponds to a technical challenge established in Phase 1 and was retained only after alternative designs were evaluated against the study's admissibility criteria.

For each video, the fixed measurement stage produces the three quantities defined in Section 3.1: a generic-safety reading, interpretable concept measurements, and frozen generic-safety evidence channels. LlavaGuard provides the generic-safety reading and visual representation used to measure semantic concepts (Helff et al. 2024), while deterministic concepts are extracted using fixed and reproducible procedures. All concept measurements are normalized using training-partition statistics and remain fixed during model training. The generic-safety evidence channels are also frozen and are used only to prevent the child-specific pathway from duplicating information already captured by generic safety. Detailed measurement definitions, scoring and extraction procedures, temporal aggregation, normalization, and fixed structural assignments are provided in Online Appendix B.1.

The risk-assessment stage maps scalar evidence through *bounded, zero-anchored, nondecreasing response functions*. Zero anchoring assigns no additional hazard at the reference value, monotonicity preserves the theoretically specified direction of risk, and boundedness limits the influence of any single signal. Piecewise responses further capture nonlinear effects such as thresholds and saturation. Generic-safety calibration uses an analogous function with separate parameters. These properties support the identifiability and faithfulness results developed below, while the exact functional forms and parameterization are provided in Online Appendix B.2.1. Throughout the methodology, the child-specific residual denotes mechanism hazards learned beyond the generic-safety channel, whereas the baseline hazard denotes input-independent risk. The following four design elements constitute the novel components of the risk-assessment architecture.

**3.2.1. Design Element 1: Identifiable Child-Specific Residual Learning**

Design Element 1 addresses Technical Challenge 1 through three linked operations: additive aggregation consistent with noisy-OR, contextual calibration of generic safety, and structural and statistical separation of child-specific residual hazards from generic safety evidence.

**Union-consistent additive risk aggregation**. Let $E_i^G$ denote the generic unsafe event, $E_i^r$ the event associated with child-specific mechanism $r$, and $E^0$ the leak event. The child-inappropriate event is $E_i = E_i^G \vee (\bigvee_{r=1}^{R} E_i^r) \vee E^0$. Let $p_i^G = P(E_i^G | X_i)$, $p_i^r = P(E_i^r | X_i)$, $p^0 = P(E^0)$, and $p_i = P(E_i | X_i)$. Under the noisy-OR modeling assumption,

$$1 - p_i = (1 - p_i^G)(1 - p^0) \prod_{r=1}^{R} (1 - p_i^r). \tag{5}$$

Defining the hazard of an event probability as $\lambda = - log(1 - p)$ gives

$$\Lambda_i = \lambda_i^G + \sum_{r=1}^{R} \lambda_i^r + \lambda^0,\ p_i = 1 - exp(-\Lambda_i). \tag{6}$$

The representation preserves the OR semantics on the probability scale while keeping all risk sources additive and nonnegative on the hazard scale. For example, if a video contains both generically unsafe violence and a child-specific imitable dangerous action, both hazards contribute to $\Lambda_i$; neither can cancel the other through a negative residual.

*Proposition 1 (Uniqueness of the noisy-OR-compatible additive link).* Let $F: R_+ \rightarrow [0, 1)$ be continuous and strictly increasing with $F(0) = 0$. If additive nonnegative risk quantities satisfy $F(a + b) = F(a) + F(b) - F(a)F(b)$ for all $a, b \geq 0$, then $F(t) = 1 - exp(-ct)$ for some $c > 0$. Hence, up to positive rescaling, the exponential link is the unique continuous strictly increasing link satisfying both additive decomposition and noisy-OR composition.

Proposition 1 shows that the exponential hazard link is determined, up to scale, by the two aggregation requirements. The proof is given in Online Appendix B.3.1.

**Context-aware calibration of the generic-safety channel.** The frozen model supplies the generic-safety reading $u_i$, whereas the additive hazard formulation requires a generic-safety hazard $\lambda_i^G$. We therefore define

$$\lambda_i^G = a_i \psi_G(u_i), \tag{7}$$

where $\psi_G(\cdot)$ is a monotone, bounded, zero-anchored calibration built from the shared response basis and $a_i \in [0, 1]$ is a contextual attenuation factor. The calibration maps the scalar reading onto the hazard scale; the attenuation factor incorporates contextual information that is not represented by the scalar generic-safety reading.

Let $E$ denote the set of *contextual attenuation concepts*, such as animation, gameplay interfaces, news, and educational or medical settings. Their normalized measurements determine

$$a_i = \prod_{e \in E} \sigma\left(\frac{\vartheta_e - \tilde{c}_{ie}}{\tau_e}\right), \tag{8}$$

where $\vartheta_e$ is fixed at the $0.90$ training-distribution quantile of contextual concept $e$, and the final model fixes the common smoothness parameter at $\tau_e = 0.25$. Stronger contextual attenuation evidence decreases $a_i$. These concepts are extracted from the same frozen visual representation as the semantic child-risk concepts but are assigned only to the generic-safety channel. The contextual thresholds and temperature are fixed measurement parameters rather than trainable model parameters.

For example, an animated fight can receive a high generic-safety reading because of visible violence. Strong animation evidence may attenuate $\lambda_i^G$ through $a_i$, but it cannot attenuate an independently supported imitable-dangerous-action mechanism hazard. The adjustment is therefore asymmetric: contextual attenuation corrects only the reused generic-safety channel and does not create a general exemption from child-specific risk. The bounded calibration and full contextual attenuation parameterization are given in Online Appendix B.2.1.

**Structural identifiability and residual-leakage control.** A faithful residual decomposition requires two complementary properties. *Structural identifiability* requires a given total hazard to have a unique component-level decomposition within the constrained model class. *Statistical separation* limits the extent to which a structurally identified child-specific mechanism reproduces the generic-safety reading.

*Proposition 2 (Identifiability of the additive residual decomposition)*. Consider two decompositions in the proposed model class that produce the same total hazard for every admissible input. If (*i*) each input-dependent component depends only on its designated non-overlapping input block, (*ii*) each such component is zero at its block-specific reference point, and (*iii*) the joint input support permits each block to vary while the others are held at their reference points, then the decompositions coincide componentwise and the input-independent baseline is uniquely assigned to $\lambda^0$.

The non-overlapping child-specific input blocks required by Proposition 2 are operationalized by the knowledge-anchored mechanism structure in Design Element 2; zero anchoring is supplied by the shared response basis above. Under the block-separability, zero-anchoring, and joint-support conditions stated in Proposition 2, the result rules out arbitrary reallocations among the generic-safety and child-specific mechanism hazards within the specified model class. The proof is given in Online Appendix B.3.2.

Structural identifiability does not imply statistical independence. Let $s_i$ collect the frozen generic-safety evidence channels for video $i$, and let $A_r = \{i: \gamma_i^r \geq \tau_{act}\}$ denote the samples on which mechanism $r$ is active. We use statistical dependence as a soft diagnostic of whether an active child-specific mechanism is statistically duplicating generic-safety evidence, and penalize it through

$$L_{sep} = \sum_{r=1}^{R} nHSIC\left(\lambda_{A_r}^{r}, S_{A_r}\right). \quad (9)$$

The frozen evidence vector $e_i$ contains auxiliary generic-safety channels used only for this separation objective; its exact composition and minibatch implementation are reported in Online Appendix B.2.4. Thus, structural identifiability provides a unique component decomposition under the conditions of Proposition 2, while the separation penalty discourages statistical duplication of generic-safety evidence without requiring strict independence.

#### 3.2.2. Design Element 2: Knowledge-Anchored Mechanism Structuring and Differentiable Gating

Design Element 2 addresses Technical Challenge 2 by determining both which evidence belongs to each child-specific mechanism and when that mechanism contributes sufficiently to the prediction.

**Knowledge-anchored mechanism structure.** We represent child-specific risk using a set of externally grounded mechanism blocks indexed by $r$. In this study, we define three child-specific mechanisms (Table 2): $r_1$: imitable dangerous actions, $r_2$: sexualized or adult-like performance, and $r_3$: self-harm or depression-related content. These mechanisms are grounded in established theories of observational learning, sexualization, and self-harm contagion, respectively, as summarized in Table 2. Importantly, the substantive boundaries of these mechanisms are determined by domain knowledge rather than learned from outcome labels. The concept assignments and gate topology described below then translate these theoretically defined mechanisms into computational representations.

For mechanism $r$, $C_r$ denotes the *mechanism concept block*: the deterministic and semantic concepts that the mechanism is allowed to use. We partition it as $C_r = C_r^{det} \cup C_r^{sem}$. Concept membership is specified before model fitting based on the externally grounded mechanism definitions and the measurement specification. The mechanism blocks are mutually disjoint:

$$C_r \cap C_s = \varnothing, \qquad r \neq s. \tag{10}$$

Each concept is assigned to only one mechanism block before model fitting. This partition prevents concept contributions from being reassigned across substantively distinct mechanisms and provides the non-overlapping input structure required by Proposition 2.

Not every concept relevant to a mechanism is sufficiently specific to activate it. We therefore organize the activation evidence of mechanism $r$ into a fixed collection of families $\{A_{rf}\}_{f=1}^{F_r}$, with $A_{rf} \subseteq C_r$. Evidence within a family is substitutable, whereas distinct families encode conjunctive requirements when substantive grounding requires multiple types of evidence to be jointly present. The concept blocks and evidence-family memberships are fixed before training; learning adjusts their quantitative responses and thresholds, not their membership.

**Differentiable mechanism gating.** The hazard of mechanism $r$ is decomposed into an activation degree $\gamma_i^r \in [0, 1]$ and a nonnegative mechanism intensity $\mu_i^r$:

$$\lambda_i^r = \gamma_i^r \mu_i^r, \qquad \mu_i^r = \mu_i^{r,det} + \mu_i^{r,sem}. \tag{11}$$

For evidence family $f$, its smooth anchor summary is

$$a_{irf} = \frac{1}{\beta_g} log\left(\sum_{k \in A_{rf}} exp(\beta_g \tilde{c}_{ik})\right), \tag{12}$$

where $\beta_g > 0$ controls the smooth maximum within an evidence family. The family response and mechanism activation are

$$z_{irf} = \sigma\left(\frac{a_{irf} - \vartheta_{rf}}{\tau_g}\right), \ \gamma_i^r = \prod_{f=1}^{F_r} z_{irf}. \tag{13}$$

Table 2 summarizes how the knowledge-grounded risk mechanisms are translated into the final ChildRiskGuard topology, distinguishing evidence that activates each mechanism gate from evidence that only modulates mechanism intensity.

Evidence-family membership remains fixed, whereas the thresholds $\vartheta_{rf}$ are learned jointly with the risk-assessment stage. The same continuous gate is used during training and inference. The selected value or schedule for $\tau_g$ is reported in Online Appendix B.2.2.

#### 3.2.3. Design Element 3: Semantic Contribution Budget

Design Element 3 mitigates the input-side faithfulness risk in Technical Challenge 3 by limiting how strongly VLM-derived semantic evidence can influence mechanism intensity. Deterministic concepts provide reproducible physical evidence, whereas semantic concepts provide contextual information unavailable from deterministic measurements. For example, skin exposure can be measured directly, while adult-like performance context requires semantic interpretation. Because a continuous VLM-derived concept score may also encode information beyond its declared semantics, we retain semantic concepts but explicitly limit their contribution through the mechanism-intensity pathway.

For mechanism $r$, define the deterministic and semantic intensity components

$$\mu_i^{r,det} = \sum_{k \in C_r^{det}} h_{rk}(\widetilde{c}_{ik}),\ \mu_i^{r,sem} = \alpha_{sem} \sum_{k \in C_r^{sem}} min\left\{h_{rk}(\widetilde{c}_{ik}), H_{sem}\right\}, \tag{14}$$

so that $\mu_i^r = \mu_i^{r,det} + \mu_i^{r,sem}$.

Here $\alpha_{sem} = 0.35$ scales the semantic branch and $H_{sem} = 2.5$ caps each semantic concept response. The *semantic contribution budget* constrains the total semantic intensity by a deterministic-dominance term plus an independent semantic slack:

$$\mu_i^{r,sem} \le \kappa_r^{dom} \mu_i^{r,det} + s_r. \tag{15}$$

The final model uses $\kappa_r^{dom} = 1$ for every mechanism and $(s_1, s_2, s_3) = (1.0, 1.0, 1.2)$. We enforce this constraint softly during training. The violation magnitude is

$$b_i^r = \left[\mu_i^{r,sem} - \kappa_r^{dom} \mu_i^{r,det} - s_r\right]_+, \tag{16}$$

and the semantic-budget penalty is the mean unsquared hinge,

$$L_{bud} = \frac{1}{D} \sum_{i=1}^{D} \sum_{r=1}^{R} b_i^r. \tag{17}$$

At inference, the final model additionally applies

$$\bar{\mu}_i^{r,sem} = min\left\{\mu_i^{r,sem}, \kappa_r^{dom} \mu_i^{r,det} + s_r\right\}, \qquad \chi_i^r = \frac{\bar{\mu}_i^{r,sem}}{max\{\mu_i^{r,sem}, \varepsilon\}}. \tag{18}$$

The per-concept cap, soft training penalty, and inference projection are all components of the primary model rather than robustness-only variants.

*Proposition 3 (Budget preservation under gating and aggregation).* Suppose Equation 15 holds for every mechanism of video $i$. Define $\beta_r = \kappa_r^{dom}/(1 + \kappa_r^{dom})$ and $\delta_r = s_r/(1 + \kappa_r^{dom})$. Let $H_i^{sem} = \sum_r \gamma_i^r \mu_i^{r,sem}$, $H_i^{child} = \sum_r \lambda_i^r$, and $\beta_{max} = max_r \beta_r$. Then

$$H_i^{sem} \le \beta_{max} H_i^{child} + \sum_{r=1}^{R} \gamma_i^r \delta_r. \tag{19}$$

When all $\delta_r = 0$ and $H_i^{child} > 0$, $H_i^{sem}/H_i^{child} \le \beta_{max}$. Thus, mechanism-level semantic contribution budgets remain meaningful after gating and aggregation rather than disappearing at the system level. The inference-time projection is specified in Online Appendix B.2.3, and the proof is provided in Online Appendix B.3.3.

The semantic contribution budget applies specifically to the mechanism-intensity pathway. If a semantic concept belongs to an activation family, its effect on mechanism activation is governed separately by the knowledge-anchored gate in Design Element 2. Contextual attenuation concepts in $E$ operate only on the generic-safety channel and are not part of this budget.

#### 3.2.4. Design Element 4: Axiomatically Faithful Monotone Additive Hazard Mapping

Design Element 4 addresses the mapping-side failure in Technical Challenge 3. The objective is to retain nonlinear concept–risk relationships while satisfying the four faithfulness axioms A1–A4 introduced earlier. We achieve this by combining the anchored monotone response functions defined above with the additive mechanism structure.

For concept $k$ in mechanism $r$, we define its realized hazard contribution as

$$\phi_{ik}^r = \{\gamma_i^r h_{rk}(\tilde{c}_{ik}),\ k \in C_r^{det},\ \gamma_i^r \chi_i^r \alpha_{sem} min\{h_{rk}(\tilde{c}_{ik}), H_{sem}\},\ k \in C_r^{sem}, \tag{20}$$

where $h_{rk}(\tilde{c}_{ik})$ captures the nonlinear risk response of the concept, $\gamma_i^r$ accounts for mechanism activation, and $\chi_i^r$ is the inference-time semantic projection factor from Equation 18; during training, $\chi_i^r = 1$. The mechanism hazard is therefore

$$\lambda_i^r = \sum_{k \in C_r} \phi_{ik}^r. \tag{21}$$

This construction preserves the flexibility of nonlinear concept responses while keeping every realized contribution explicit. For example, skin exposure may have little effect within a benign range, increase after reaching a risk-relevant level, and eventually saturate; its contribution to the final hazard remains directly observable through $\phi^{r}_{ik}$ rather than being hidden inside an unconstrained nonlinear prediction head.

*Proposition 4 (Axiomatic faithfulness of the child-specific pathway).* Under the nonnegative additive hazard aggregation of Design Element 1, the disjoint knowledge-anchored mechanism structure of Design Element 2, and the anchored monotone additive mapping above, the child-specific prediction pathway satisfies A1–A4. Proposition 4 establishes that the four faithfulness requirements are structural properties of the artifact rather than post-hoc characteristics measured after training. The formal definitions of A1–A4 and the complete verification are provided in Online Appendix B.3.4.

When the aggregate semantic projection is inactive ($\chi^{r}_{i} = 1$), the additive mapping further yields an exact attribution result: conditional on a realized mechanism gate $\gamma^{r}_{i}$, the concept contribution $\phi^{r}_{ik}$ coincides with the Shapley value of concept $k$ in the corresponding additive intensity game. When projection is active, exact reconstruction and monotonicity remain valid, but we do not claim unconditional Shapley equivalence. The additive-regime result is stated and proved in Online Appendix B.3.5.

The four design elements form a single risk-assessment function and are optimized jointly rather than as separate predictors. The measurement stage remains frozen: LlavaGuard, the shared vision encoder, the Contrastive Language–Image Pretraining (CLIP) text encoder (Radford et al. 2021), and the deterministic concept extractors receive no gradient. Trainable parameters are confined to the risk-assessment stage, including the generic-safety calibration, mechanism-gate thresholds, concept-response functions, and the input-independent baseline hazard; contextual attenuation thresholds remain fixed. Thus, improvements in child-specific prediction are attributable to the proposed risk-assessment architecture rather than adaptation of the underlying measurement models.

For training, the total hazard $\Lambda_i$ defined in Equation (2) is converted to a child-inappropriateness probability through the exponential hazard link, and this probability enters the class-weighted binary cross-entropy loss. The four design elements are jointly estimated using

$$L = L_{BCE} + \xi_{sp} L_{sparse} + \xi_{bud} L_{bud} + \xi_{sep} L_{sep}, \tag{22}$$

where $L_{BCE}$ is the class-weighted binary cross-entropy loss; $L_{sparse}$ regularizes the realized child-specific mechanism hazards rather than selecting a discrete subset of concepts; $L_{bud}$ is the semantic-budget penalty defined in Design Element 3; and $L_{sep}$ is the residual-separation penalty defined in Design Element 1. The coefficients $\xi_{sp}$, $\xi_{bud}$, and $\xi_{sep}$ control their respective contributions. The training objective, residual-separation penalty, probability calibration, and decision rule are specified in Online Appendix B.2.4. This objective is not merely an optimization device: its regularization terms operationalize the structural requirements introduced by the corresponding design elements.

The same hazard-to-probability mapping is used during training and inference. For total hazard $\Lambda_i = \lambda_i^G + \sum_{r=1}^{R} \lambda_{ir} + \lambda_0$, the predicted probability of child-inappropriateness is

$$\hat{p}_i = 1 - exp\left(-\Lambda_i\right). \tag{23}$$

where $\lambda_i^G$ is the calibrated generic-safety hazard, $\lambda_{ir}$ is the hazard contributed by child-specific mechanism $r$, and $\lambda_0$ is the input-independent baseline hazard. The moderation decision compares $p_i$ with an operating threshold selected exclusively from five-fold out-of-fold predictions on the development data; this threshold is locked before evaluation on the test set.

To report how the total predicted hazard is distributed across its sources, we additionally define normalized hazard shares. For $\Lambda_i > 0$,

$$\pi_i^G = \frac{\lambda_i^G}{\Lambda_i}, \qquad \pi_i^r = \frac{\lambda_i^r}{\Lambda_i}, \qquad \pi_i^0 = \frac{\lambda^0}{\Lambda_i}. \tag{24}$$

where $\pi_i^G$, $\pi_{ir}$, and $\pi_i^0$ denote the proportions of total hazard attributable to generic safety, child-specific mechanism $r$, and baseline risk, respectively; these shares sum to one. When $\Lambda_i = 0$, all component hazards are zero and we set the corresponding shares to zero by convention. Together with the concept contributions defined in Design Element 4, the same forward pass therefore yields the prediction and its generic-safety, mechanism-level, and concept-level decomposition, without requiring a separate post-hoc explanation model.

### 3.3. Design Novelty of ChildRiskGuard within AI4CDS

ChildRiskGuard constitutes the artifact-level outcome of the Phase 2 resource construction and Phase 3 design search described by AI4CDS. The four retained design elements jointly address the three technical challenges established in Section 2.4: identifiable residual learning separates generic from child-specific risk; knowledge-anchored mechanism structuring represents heterogeneous developmental risks; the semantic contribution budget constrains input-side faithfulness risks; and the monotone additive mapping preserves identifiable concept-level attribution. Importantly, these elements represent the retained outcome of a broader design search rather than a design specified ex ante. Online Appendix A.3 documents the alternative architectures, mechanisms, constraints, and response functions that were accepted, modified, reclassified, or rejected before the final ChildRiskGuard specification was reached.

## 4. Evaluations

This section implements Phase 4 of AI4CDS: AI-Enabled Multi-Faceted Evaluation. This phase (1) fixes evaluation rules in advance, with fitting and threshold selection confined to development data before final test evaluation; (2) changes one focal quantity at a time in AI-supported ablations and diagnostics; (3) reconciles AI-supported implementation with the artifact specification, correcting code, description, or claims when they diverge; (4) compares benchmarks under common development-selected operating rules; and (5) treats AI-generated interpretations as proposals that researchers verify against conflicting evidence and anomalies. In ChildRiskGuard, these checks yield evidence on predictive utility, design-element and mechanism contributions, explanation faithfulness, robustness, and generalizability. Detailed specifications, corrections, and audit records are articulated in Online Appendix A.4.

### 4.1. Data and Evaluation Design

We evaluate ChildRiskGuard on 7,070 short-form videos: 1,000 child-inappropriate videos from the expert-annotated TikGuard dataset (Balat et al. 2024) and 6,070 additional TikTok videos manually reviewed and labeled by a trained coauthor serving as the study annotator, following the same child-appropriateness criteria. Stratified sampling produces a development set of 6,010 videos and an untouched test set of 1,060 videos, including 150 positive cases. The development set is used for model fitting and five-fold out-of-fold threshold selection; the test set is accessed only for final evaluation.

We compare ChildRiskGuard with four benchmark families representing increasingly flexible alternatives: conventional classifiers using the same frozen measurements, concept-bottleneck models, end-to-end video classifiers, and large vision-language or language models. These comparisons test whether the proposed architecture adds value beyond its measurements, conventional concept-based prediction, task-specific representation learning, and general-purpose multimodal capabilities. The frozen LlavaGuard baseline isolates the generic-safety capability reused by ChildRiskGuard, while task-specific fine-tuning tests whether direct adaptation to the same child-appropriateness labels is sufficient without structured risk decomposition.

We report precision, recall, and F1 at the selected operating point, using F1 as the primary comparison metric because moderation ultimately requires a referral decision. Learned models are evaluated across five random seeds and reported as mean ± standard deviation.

**4.2. ChildRiskGuard Prediction Utility**

Table 3 compares ChildRiskGuard across the four benchmark families. (1) Against conventional classifiers using the same frozen measurements (Table 3 Panel A), ChildRiskGuard achieves F1 = 0.769 versus 0.593 for the strongest baseline, Random Forest. This difference indicates that predictive utility cannot be attributed to the measurement representation alone; how the measurements are structured and combined also matters. (2) We compare ChildRiskGuard with CBMs to assess whether a conventional concept-based framework is sufficient for this task. As shown in Table 3 (Panel B), ChildRiskGuard achieves an F1 score of 0.769, outperforming the strongest CBM baseline, LM4CV, at 0.758. PCBM achieves higher recall but lower precision, indicating a different precision-recall trade-off rather than consistently better performance. Because all models use the same child-appropriateness labels, the difference lies not in whether they consider the target audience, but in how they model child-specific risk. ChildRiskGuard distinguishes generic from child-specific risks, organizes the latter around externally grounded developmental mechanisms, and incorporates the structural faithfulness requirements described in Section 3. (3) Table 3 (Panel C) examines whether end-to-end deep learning models can learn child appropriateness without an explicit, domain-structured representation of risk. ChildRiskGuard achieves an F1 score of 0.769, outperforming the strongest benchmark, the frame-wise ResNet-18 classifier (F1=0.711). This result highlights the value of incorporating domain-specific requirements into the predictive architecture rather than relying solely on task-specific representation learning. Panel C also includes two recent IS models, KG-NTM (Xie et al. 2026) and Agenda-setting BERT (Kim et al. 2025), both developed for contextualized video classification. Because these models were designed for different inputs and objectives, their results serve as supplementary comparisons rather than directly comparable benchmarks. (4) Table 3 (Panel D) evaluates whether ChildRiskGuard's contribution can be explained by general-purpose vision-language capabilities, prompt engineering, or task-specific fine-tuning. The frozen LlavaGuard baseline is particularly informative because it provides the same generic-safety capability reused within ChildRiskGuard. ChildRiskGuard raises F1 from 0.632 to 0.769, while InternVL3 reaches 0.758 and fine-tuned LlavaGuard reaches 0.679. These results do not support a claim of universal predictive superiority over foundation models; rather, they show that ChildRiskGuard remains competitive while providing structured attribution at the generic-safety, mechanism, and concept levels within the same predictive computation. The compared prompting and fine-tuning baselines do not provide this decomposition, and ChildRiskGuard obtains it without sample-level mechanism annotations.

**Table 3. Predictive Performance Across Benchmark Families**

| **Method** | **F1** | **Precision** | **Recall** |
|---|---|---|---|
| ChildRiskGuard (Ours) | **0.7689 ± 0.0019** | **0.7940 ± 0.0007** | 0.7453 ± 0.0030 |
| *Panel A: Conventional Machine Learning* | | | |
| Logistic Regression | 0.5396 ± 0.0000 | 0.4817 ± 0.0000 | 0.6133 ± 0.0000 |
| Random Forest | 0.5928 ± 0.0064 | 0.5722 ± 0.0213 | 0.6160 ± 0.0209 |
| XGBoost | 0.5851 ± 0.0102 | 0.5668 ± 0.0538 | 0.6107 ± 0.0379 |
| Tabular Multilayer Perceptron | 0.5894 ± 0.0147 | 0.6161 ± 0.0741 | 0.5760 ± 0.0616 |
| *Panel B: Concept Bottleneck Models* | | | |
| Label-free CBM (Oikarinen et al. 2022) | 0.7281 ± 0.0055 | 0.6790 ± 0.0339 | 0.7880 ± 0.0381 |
| PCBM (Yuksekgonul et al. 2022) | 0.7565 ± 0.0019 | 0.7230 ± 0.0047 | 0.7933 ± 0.0082 |

| Concept-Aligned LaBo (Yang et al. 2023) | 0.7359 ± 0.0351 | 0.6769 ± 0.0394 | 0.8080 ± 0.0506 |
|---|---|---|---|
| LM4CV (Yan et al. 2023) | 0.7577 ± 0.0208 | 0.7819 ± 0.0329 | 0.7373 ± 0.0444 |
| *Panel C: Deep Learning Methods* | | | |
| CNN (ResNet-18) | 0.7114 ± 0.0251 | 0.7001 ± 0.0555 | 0.7280 ± 0.0451 |
| RNN (ResNet-18 + GRU) | 0.7017 ± 0.0209 | 0.6955 ± 0.0269 | 0.7093 ± 0.0373 |
| 3D CNN (R3D-18) | 0.6503 ± 0.0110 | 0.6381 ± 0.0333 | 0.6653 ± 0.0318 |
| KG-NTM (Xie et al. 2026) | 0.6263 ± 0.0573 | 0.6430 ± 0.0487 | 0.6147 ± 0.0865 |
| Agenda-setting BERT (Kim et al. 2025) | 0.3569 ± 0.0090 | 0.2452 ± 0.0051 | 0.6587 ± 0.0504 |
| *Panel D: Large Language and Vision-Language Methods* | | | |
| LlavaGuard judge only (Helff et al. 2024) | 0.6318 ± 0.0199 | 0.5572 ± 0.0486 | 0.7360 ± 0.0396 |
| Qwen3-VL zero-shot (Bai et al. 2025) | 0.7159 ± 0.0041 | 0.7017 ± 0.0034 | 0.7307 ± 0.0076 |
| InternVL3 zero-shot (Zhu et al. 2025) | 0.7579 ± 0.0067 | 0.6869 ± 0.0060 | **0.8453 ± 0.0110** |
| Expert-Augmented Prompt Learning (Wei et al. 2025) | 0.2914 ± 0.0171 | 0.1878 ± 0.0214 | 0.6853 ± 0.1350 |
| GPT-5.6 Luna Zero-shot (OpenAI 2026) | 0.5733 ± 0.0118 | 0.6293 ± 0.0172 | 0.5267 ± 0.0125 |
| GPT-5.6 Luna Few-shot In-context-learning | 0.6824 ± 0.0083 | 0.6115 ± 0.0085 | 0.7720 ± 0.0119 |
| GPT-5.6 Luna Chain-of-thought | 0.3887 ± 0.0130 | 0.7170 ± 0.0211 | 0.2667 ± 0.0105 |
| GPT-5.6 Luna Self-refinement | 0.3990 ± 0.0202 | 0.6439 ± 0.0183 | 0.2893 ± 0.0192 |
| GPT-5.6 Luna Prediction-feedback | 0.6475 ± 0.0120 | 0.5445 ± 0.0079 | 0.7987 ± 0.0218 |
| Fine-tuned LlavaGuard (Helff et al. 2024) | 0.6793 ± 0.0092 | 0.6052 ± 0.0181 | 0.7747 ± 0.0110 |

### 4.3. Ablation, Sensitivity, and Generalizability Analyses

Table 4 examines the sources of ChildRiskGuard's predictive gains through component- and mechanism-level ablations. Panel A removes the focal innovation of each design element while holding the rest of the architecture constant. All four innovations improve performance. The largest F1 gains come from the calibrated generic channel in Design Element 1 and the monotone additive experts in Design Element 4, with increases of 0.085 and 0.077, respectively. The semantic budget in Design Element 3 adds 0.012, while mechanism gating in Design Element 2 adds 0.006. Panel B shows that including M1, M2, and M3 increases F1 by 0.035, 0.011, and 0.150, respectively, with the self-harm and depressive-risk mechanism contributing the largest incremental gain. These ablations show that M1–M3 make non-redundant predictive contributions in the focal dataset. Online Appendix C.1 shows that the architecture can accommodate additional grounded mechanisms: M4 is added using mechanism-specific measurements without redesigning the core risk-assessment architecture, modifying M1–M3, or fine-tuning the frozen foundation model.

**Table 4. Ablation Studies**

| **Method** | **F1** | **Precision** | **Recall** |
|---|---|---|---|
| *Panel A. Component-level Ablation of Key Design Innovations* | | | |
| ChildRiskGuard (Ours) | **0.7689 ± 0.0019** | **0.7940 ± 0.0007** | **0.7453 ± 0.0030** |
| w/o Design Element 1, calibrated generic channel | 0.6838 ± 0.0010 | 0.6457 ± 0.0017 | 0.7267 ± 0.0000 |
| w/o Design Element 2, mechanism gating | 0.7627 ± 0.0004 | 0.7853 ± 0.0042 | 0.7413 ± 0.0030 |
| w/o Design Element 3, semantic budget | 0.7569 ± 0.0019 | 0.7805 ± 0.0007 | 0.7347 ± 0.0030 |
| w/o Design Element 4, monotone additive experts | 0.6921 ± 0.0469 | 0.6837 ± 0.0680 | 0.7027 ± 0.0361 |
| *Panel B. Mechanism-level ablation* | | | |
| ChildRiskGuard (Ours) | **0.7689 ± 0.0019** | **0.7940 ± 0.0007** | 0.7453 ± 0.0030 |
| w/o M1, dangerous action | 0.7340 ± 0.0019 | 0.7008 ± 0.0105 | 0.7707 ± 0.0089 |
| w/o M2, adultized performance | 0.7576 ± 0.0055 | 0.7418 ± 0.0203 | **0.7747 ± 0.0110** |
| w/o M3, self-harm and depressive risk | 0.6185 ± 0.0019 | 0.6145 ± 0.0017 | 0.6227 ± 0.0037 |

We assess sensitivity to the semantic representation by replacing the semantic readout encoder while holding the remaining architecture fixed; these results are reported in Online Appendix C.2.

### 4.4. Interpretable Insights: Faithfulness and Mechanism-Level Interpretation

ChildRiskGuard requires only the final binary risk label (e.g., 0 or 1), for each video during training. It does not require additional labels for individual mechanisms or concepts. Despite this limited supervision, ChildRiskGuard can still provide mechanism-level and concept-level attributions. This is because these contributions are computed directly within the same forward pass that generates the prediction, and together they reconstruct the child-specific risk score. The following cases provide complementary instance-level evidence by examining whether these mechanism- and concept-level contributions are decision-relevant in concrete test examples. We present one representative case for each mechanism in the primary M1–M3 instantiation and report the generic and child-specific hazards, concept-level contributions, and the prediction after removing the focal mechanism.

In Figure 2(a), the video contains a dangerous fire-performance activity (M1). LlavaGuard assigns an unsafe probability of 0.221 and classifies it as safe, whereas ChildRiskGuard predicts a child-inappropriate probability of 0.738, above the 0.612 threshold. M1 is mainly supported by the risky-action semantic concept, which contributes 0.702 hazard, together with motion evidence from global flow (0.036), flow peak (0.036), motion energy (0.031), and body motion (0.004). At the channel level, M1 contributes 0.809 hazard, compared with 0.520 from the generic channel, while M2 and M3 contribute almost no risk. Removing M1 reduces the predicted probability from 0.738 to 0.412 and changes the decision from child-inappropriate to safe, showing that M1 is decision-critical in this example.

Figure 2(b) shows skin exposure in an adultized performance context (M2). LlavaGuard assigns an unsafe probability of 0.138 and classifies the video as safe, whereas ChildRiskGuard assigns 0.707. The M2 explanation identifies multiple context-specific signals, including nightclub dance (0.109), sexualized dance (0.074), revealing outfit (0.036), suggestive pose (0.021), suggestive context (0.011), and adult styling (0.008), together with a deterministic skin-exposure contribution of 0.028. These results show that M2 identifies child-inappropriate content by combining physical exposure with semantic evidence of an adultized or sexualized performance context, rather than relying on exposed skin alone. The generic and M2 channels contribute hazards of 0.938 and 0.286, respectively, while M1 and M3 remain negligible. Removing M2 lowers the probability from 0.707 to 0.610, below the operating threshold, thereby reversing the final decision.

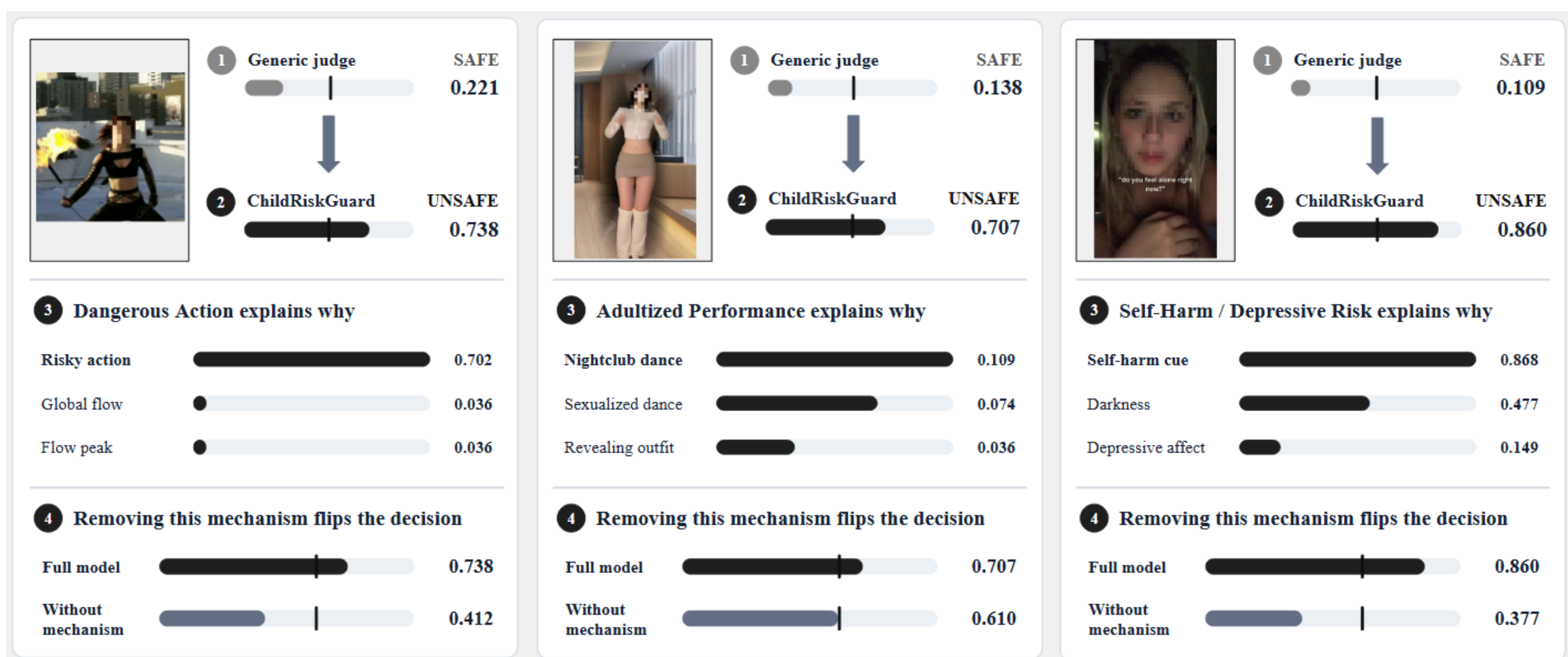

**Figure 2. ChildRiskGuard: Mechanism-Level Explanations**

Figure 2(c) shows a dark visual scene with self-harm or depressive evidence (M3). LlavaGuard assigns an unsafe probability of 0.109 and classifies the video as safe, whereas ChildRiskGuard assigns 0.860. M3 is supported by the self-harm semantic concept (0.868), deterministic darkness measurement (0.477), and depressive affect (0.149). At the channel level, M3 contributes a hazard score of 1.493, substantially higher than the generic-channel hazard of 0.454, while M1 and M2 contribute almost nothing. Removing M3 reduces the predicted probability from 0.860 to 0.377 and reverses the classification. This case shows that the child-specific residual channel can detect implicit self-harm and depressive risks missed by the generic safety judge.

Across all three cases, removing the focal mechanism reverses the final decision, confirming that the reported explanations reflect the model's actual decision process. This model-internal causal relationship should not be interpreted as evidence that the corresponding developmental mechanisms cause real-world harm, whose substantive interpretation relies on the external domain grounding in Sections 2 and 3.

AI support in Phase 4 did not uniformly improve the evaluation. It also generated invalid analyses and premature interpretations: an operating point informed by test data, an abstention rule that withheld too many decisions, a misdescription of the M2 gate, an inadequately qualified component contribution, and a recommendation to remove a mechanism based on one split. Researchers removed, revised, corrected, or rejected these outputs after review (Online Appendix A.4, Table A3). These incidents show that AI can expand evaluation and diagnosis while also creating verification burdens when computational outputs become scientific evidence or claims.

## 5. Discussion and Conclusion

This section implements Phase 5 of AI4CDS: AI-Assisted Design Principle Distillation. This phase (1) assembles the cross-phase record of requirements, decisions, evaluation evidence, corrections, and interventions; (2) traces how ChildRiskGuard's Phase 1 requirements constrained later resource and design choices and how Phase 4 evidence supported, revised, or qualified them; and (3) uses AI to surface recurring patterns. Researchers then (4) test candidate abstractions against the evidence, case dependence, and contradictions, narrowing or rejecting claims that exceed support, and (5) distill surviving patterns into transferable knowledge grounded in theory. This process yields methodological implications, constitutive-use evidence, governance principles, reproducibility guidance, domain implications, and boundary conditions. Detailed records and procedures are articulated in Online Appendix A.5.

### 5.1. AI4CDS as a Prescriptive Methodological Framework

AI4CDS is a prescriptive methodological framework for conducting CDS when AI participates across the research lifecycle. Building on human–AI delegation research, AI4CDS extends the delegation problem from task allocation to task- and phase-contingent AI discretion across problem formulation, resource construction, design search, evaluation, and knowledge abstraction (Baird and Maruping 2021,

Fügener et al. 2022). Across these phases, AI expands candidate generation and analysis, while researchers retain authority over domain grounding, admissibility, verification, and scientific commitments. AI4CDS also makes domain requirements traceable across the research process. Domain knowledge helps define the consequential problem and technical challenges, constrains the resources and designs considered admissible, and determines the evidence required for evaluation. The framework therefore treats AI-enabled CDS as a governed research process in which AI expands scientific search but does not determine scientific commitments. Its contribution is to make these roles, transitions, and responsibilities explicit so that AI-enabled CDS processes can be examined, documented, and refined across future studies.

### 5.2. Constitutive AI in CDS: Breadth, Depth, and Artifact Development

AI participation becomes constitutive in AI4CDS when it materially expands what can be searched, tested, revised, and completed within the research cycle. In ChildRiskGuard, AI participated across all five phases of artifact development. Table 5 combines the documented research record with literature-calibrated estimates of human workload within the same approximately time window, showing how AI increased the breadth of scientific search, the depth of iterative inquiry, and the scope of artifact development that could be completed within fixed time and researcher constraints.

**Table 5. Constitutive AI Participation Across the Five AI4CDS Phases**

| **Constitutive Dimension / AI4CDS Phase** | **With AI: Documented[a]** | **Without AI: Estimated for the Same Time Window[b]** |
|---|---|---|
| **Breadth of knowledge search**<br>Phase 1: AI-Expanded Research Problem Formulation | 7 recurring review operations; 9 computational method categories across two literatures; synthesis yielded 2 higher-order gaps and 3 technical challenges. | Systematic reviews commonly require about 12 months or longer; rapid reviews have a median duration of about 3 months and typically narrow scope or procedures (Ganann et al. 2010, Abou-Setta et al. 2016). A human-only three-month project would likely require a substantially narrower evidence synthesis[c]. |
| **Depth of resource search**<br>Phase 2: AI-Augmented Data & Resource Construction | Candidate representations, concepts, extractors, prompts, and safety resources were examined through 4 validity checks; later diagnostics repeatedly reopened resource and measurement choices. | Model construction accounts for only about 14% of typical machine learning project effort, with most work occurring in preparation and post-processing (Munson 2012). Human-only resource exploration would therefore compete directly with time available for design and evaluation. |
| **Breadth of design search** | 23 consequential design decisions, including 15 AI or joint-origin decisions, spanning model families, | Experts average 5.5 hours to formulate one research idea, while execution of one assigned idea can exceed 100 hours (Si et al. 2025, 2026). Given the full project |

| Phase 3:<br>AI-Generative Research Design Search & Refinement | mechanisms, objectives, gates, constraints, and response functions; 4 design elements were retained. | workload, approximately 3–5 major alternatives could reasonably be developed to comparable evaluable depth. |
|---|---|---|
| **Depth of iteration and diagnosis**<br>Phase 4: AI-Enabled Multi-Faceted Evaluation | 23 comparator methods across 4 benchmark families, 7 primary ablations, 5 random seeds, robustness analyses, and 8 documented correction, rejection, qualification, or verification cases. | Reproducing others' computational results can demand substantial human effort: surveyed researchers reported willingness to invest about 23 hours on average, while an intensive reproduction study allocated up to 40 human hours per paper (Elmenreich et al. 2019, Krafczyk et al. 2021). A realistic human-only scope is approximately 6–10 core benchmarks and 2–4 major diagnostic cycles. |
| **Scope of artifact development and abstraction**<br>Phase 5: AI-Assisted Design Principle Distillation | Cross-phase synthesis supported 4 retained design elements, 4 formal propositions, and the theory-informed abstraction and instantiation of 3 governance principles, differentiated reproducibility guidance, boundary conditions, and an auditable decision record. | No numeric estimate. Published workload evidence does not support a credible conversion from researcher time to propositions or principles. Within the same deadline, greater effort here would reduce the time available for Phases 1–4, and vice versa. |

**Note.** [a] "With AI" reports observed evidence from the ChildRiskGuard research record and Online Appendix A.1–A.5.
[b] "Without AI" estimates assume the same researchers, problem, data access, and effective research window, without AI support or additional personnel. The estimates of 3–5 major alternatives, 6–10 core benchmarks, and 2–4 diagnostic cycles are literature-calibrated scenario estimates based on reported task-level effort.
[c] The approximately three-month comparison window reflects the effective research period available for this study, not an assumed productivity horizon. The authors initiated the project in response to the Special Issue call for papers received in May 2026, with submission due September 7, 2026. Allowing for project initiation and final submission preparation, approximately three months were available for substantive research and artifact development.

Table 5 shows why AI made a *substantive and consequential contribution* to this research process. Within the same research window, AI participation expanded cross-domain synthesis, resource exploration, design search, benchmarking, diagnosis, and revision sufficiently to change the feasible scope and trajectory of artifact development. A human-only team could plausibly have produced a computational artifact, but matching the documented breadth and depth would likely have required a longer timeline, additional personnel, or a substantially narrower inquiry. The resulting research process was *qualitatively different because GenAI materially expanded what could be investigated, developed, and evaluated* within the same research window. It broadened knowledge search, design alternatives, implementation, diagnosis, and cross-phase synthesis. Across the research process, AI-generated outputs required verification, revision, or rejection before becoming scientific commitments. The collaboration was therefore most valuable when GenAI expanded the scope and depth of inquiry while researchers retained authority over scientific commitments.

### 5.3. Governance Principles for AI-Enabled CDS

Constitutive AI creates three governance risks in CDS: uncertain reliance, path-dependent commitment, and weakened provenance and accountability. AI4CDS governs consequential transitions at which a provisional AI-supported output becomes or changes a scientific commitment, such as an admitted resource, retained design, accepted result, or claim. Human scientific authority remains nondelegable, although AI discretion may vary. Drawing on trust calibration (Lee and See 2004, Fügener et al. 2022), design fixation and iterative design (Jansson and Smith 1991, Hevner et al. 2004), and accountability and provenance (Nissenbaum 1996, Gopal et al. 2025), AI4CDS derives three prescriptive principles (Table 6).

**Table 6. Governance Principles for AI-Enabled Computational Design Science**

| Principle | Operational Rule | ChildRiskGuard Evidence |
|---|---|---|
| **Graduated Trust** | Calibrate AI discretion to task constraints, independent verifiability, and error consequence. Constrained, verifiable tasks permit greater discretion; open-ended or consequential tasks require human control. Delegation is task-contingent, not phase-monotonic. | Online Appendix A.3/Table A2 shows AI-generated candidates under researcher-defined admissibility; A.4/Table A3 shows implementation/diagnostic support under fixed specifications, with outputs corrected or rejected before commitment. |
| **Reversibility** | When later evidence challenges an earlier commitment, reopen the earliest implicated commitment and retain, revise, replace, or reject it. Downstream modification cannot substitute for reassessing a contradicted upstream commitment. | Table A2 records mechanism reductions, dangerous-objects rejection, and gate revisions; Table A3 records a proposed mechanism removal rejected when broader evidence did not support it. |
| **Auditability** | At each consequential transition, preserve a consequence-proportional record of context, AI contribution, human disposition, decision basis, and downstream scientific consequence. | Table A2 records candidate origin, disposition, basis, and consequence; Table A3 records evaluation issues and researcher dispositions; A.5 links consequential decisions to evidence before abstraction. |

Auditability also requires differentiated reproducibility: fixed procedures should be computationally reproducible; stochastic computations statistically reproducible; AI-assisted search, diagnosis, and abstraction procedurally reconstructable; and formal claims independently verifiable from assumptions and derivations. Online Appendix A.1–A.5 supplies the process record; Online Appendix B.3 supplies formal verification. These are different forms of rigor, not weaker levels. Accumulated confidence cannot override evidence triggering reversibility; consequential reversals must remain auditable; governance effort should scale with scientific consequence and propagation risk. Appendix D provides full derivation and operationalization.

### 5.4. ChildRiskGuard as Consequential Instantiation and Domain Implications

ChildRiskGuard provides consequential feasibility evidence for AI4CDS in a setting where domain knowledge changes what constitutes an acceptable computational solution. The study translates audience-dependent safety and explanation faithfulness into technical challenges that guide resource construction, design selection, and evaluation. The resulting artifact separates generic safety from child-specific risk, represents distinct developmental-risk mechanisms, limits the influence of semantic features that may encode the target label too directly, and derives concept-level explanations from the predictive computation. Its evaluation shows a substantial improvement over the reused generic-safety model and competitive performance relative to strong foundation-model alternatives. The contribution therefore lies in a domain-grounded design combining competitive prediction with structured and faithful attribution, without claiming universal predictive superiority. The artifact also has implications for short-form video safety. Child appropriateness should be assessed separately from generic content safety, and mechanism-level decomposition can support review, escalation, and auditing by showing whether risk is associated with imitable dangerous actions, adultized performance, or self-harm and depressive evidence. These outputs should support human judgment rather than determine policy or intervention independently, because the model estimates computationally defined risks and does not establish causal harm to individual children.

**5.5. Boundary Conditions and Future CDS Research**

AI4CDS is supported by one intensive computational design project and should therefore be treated as a prescriptive framework that requires further examination across settings. The study does not establish that its five phases are the only effective organization of AI-enabled CDS, or that AI4CDS reduces research effort, expands search more effectively than alternative workflows, or produces superior artifacts. It also does not determine whether different sequences of AI interactions would have produced the same design. Future research should compare AI4CDS with alternative human-AI and human-only processes across CDS domains, examining how AI affects the alternatives considered, where researcher intervention matters most, when graduated trust, reversibility, and auditability improve research quality, and which differentiated verification standards are appropriate for different AI-enabled research activities.

ChildRiskGuard has narrower artifact-level boundaries. Its evaluation concerns one short-form video setting and a bounded set of developmental-risk mechanisms. Our binary labels do not permit estimating M1–M3's coverage of all child-specific risks. The additional mechanism analysis shows that the architecture can accommodate another grounded mechanism, but does not establish comprehensive coverage or generalization across platforms, languages, cultures, or age groups. Performance also varies with the semantic representation used, indicating dependence on the underlying pretrained measurement resources. Future work should therefore test cross-platform and temporal generalization, extend the mechanism inventory when substantive knowledge and reliable measurements support it, and evaluate the artifact with relevant stakeholders. More broadly, AI4CDS shifts the methodological question from whether AI can assist individual research tasks to how AI participation should be organized, governed, and evaluated when its outputs shape subsequent scientific decisions and the design knowledge produced from them.

# AI for Computational Design Science: A Responsible Human-AI Framework and Case Study on Short-Form Video Safety Surveillance

## Online Appendix

## Appendix A AI for Computational Design Science (AI4CDS): Process Documentation and Audit Trail

This appendix explains how generative artificial intelligence (AI) was used in the research process that produced ChildRiskGuard. It complements the technical specification in Appendix B, which defines the final measurement stage, risk assessment architecture, training objective, theoretical properties, and training and inference procedure. Here, we describe how AI support and human judgment were combined in literature search, design development, resource validation, implementation, and verification.

The research used two paid, general-purpose generative AI services between May and September 2026: ChatGPT Plus (OpenAI) and Claude Pro (Anthropic). Throughout the study, we selected the highest-capability reasoning model available to our subscription at the time of use. For ChatGPT Plus, this included GPT-5.5 Thinking during the earlier part of the study and GPT-5.6 Sol after it became available to our account. For Claude Pro, the corresponding models progressed from Claude Opus 4.7 to Claude Opus 4.8 and subsequently Claude Opus 5 as these models became available during the study period. Because these consumer services were updated during the project, the specific model used depended on the date of the interaction rather than being fixed for the entire study. Both systems were accessed through their paid web-based and desktop application interfaces; no API-based fine-tuning of these GenAI systems was performed. The models were used for the research activities documented in Appendix A.1–A.5, including literature retrieval and synthesis, resource and design exploration, implementation and diagnostic support, critical review, and cross-phase synthesis. All consequential outputs remained subject to the researcher verification and authorization procedures described below.

The framework assigns distinct responsibilities to AI and the researchers. AI was used mainly to broaden literature retrieval and design search and to support implementation and diagnostic analysis. The researchers remained responsible for defining the problem, deciding which sources and designs were admissible, assessing domain validity, setting the evaluation protocol, protecting research integrity, and interpreting the results. The five sections below correspond directly to Phase 1 through Phase 5 in the main text. Each section documents the work completed in that phase and the respective roles of AI and the researchers.

### A.1 Phase 1: AI-Expanded Research Problem Formulation

Phase 1 used the literature review to formulate the consequential research problem and derive the technical challenges that guided the artifact design. The review covered two forms of knowledge. The descriptive stream examined child development, media safety for children, regulation, and platform practice. The prescriptive stream examined inappropriate video detection, interpretable prediction, concept bottleneck models, and related computational approaches.

AI mainly supported retrieval and synthesis, especially in the larger prescriptive stream. The researchers verified the sources and assessed what the evidence implied for the research setting. After the literature review had established the relevant evidence, the researchers led the definition of the consequential problem, the derivation of domain requirements, and the formulation of the technical challenges. AI supported these decisions by organizing evidence, questioning draft formulations, and discussing alternative interpretations. The researchers made the final decisions about scope and wording.

The procedure involved seven recurring operations. New evidence or later design decisions could lead the researchers to revisit an earlier operation.

1. **Set the initial review scope**. The review began with the broad concern of short videos that may be inappropriate for children. This initial scope guided the search without fixing the final research problem in advance.
2. **Review descriptive and prescriptive knowledge**. The descriptive stream examined how risks to children differ from general safety concerns and how platforms and regulators respond to them. The prescriptive stream examined the formulation of inappropriate video detection and the available approaches to interpretable prediction.
3. **Retrieve and organize the prescriptive literature**. AI helped identify field terminology, benchmark names, and related method families. This vocabulary informed queries to the bibliographic sources used in the study. Citation expansion and searches for related work broadened the candidate set. Bibliographic metadata supplied information about publication venues and status.

4. **Verify and synthesize the evidence**. AI provided provisional relevance assessments and structured summaries. The researchers checked the selected references, reviewed the claims attributed to them, and decided which records were relevant. The synthesis documented the problem addressed, model form, supervision, inspectable quantities, stated limitations, relevant population, and evidentiary basis.
5. **Define the consequential problem**. Drawing on the reviewed evidence, the researchers defined the problem as the detection of short videos that are inappropriate for children. They also specified who would use the output and how the output could support moderation or guardianship decisions. AI was used as a discussion partner when the scope or implications of the problem remained unclear.
6. **Derive domain requirements and technical challenges**. Evidence from child development showed that appropriateness depends on the audience and that risks to children arise through different developmental mechanisms. The prescriptive literature showed that many existing explanations are produced separately from the prediction. The researchers used these findings to formulate three technical challenges. The first was to preserve useful general safety capability while identifying additional risks to children. The second was to represent heterogeneous developmental risk mechanisms. The third was to keep concept level explanations faithful to the evidence entering the model and to the computation producing the prediction. AI helped test the clarity and internal consistency of this mapping, while the researchers approved the final formulation.
7. **Return to the literature when needed**. Unsupported gap claims, new domain requirements, and changes in the candidate base method prompted further retrieval. The review expanded from detection methods to interpretable video models and concept bottleneck models. It later included concept leakage when the design moved toward a concept bottleneck formulation that did not require concept labels.

Phase 1 produced the consequential problem definition, the two literature gaps reported in Section 2, the associated domain requirements, and the three technical challenges that guide Phase 2 resource construction and Phase 3 design search.

The following prompts illustrate how AI was used in this phase. They show the main task constraints given to the model. Variable topic lists, source text, and fields outside the task are omitted for readability.

**Prompt A: Vocabulary discovery.** *For each research topic, identify recent surveys, highly cited core papers, and benchmark or dataset papers. From their titles and abstracts, extract the technical terms used by researchers, benchmark dataset names, and active research groups. Preserve the terminology used in the sources. Do not translate it into the wording of our initial problem description.*

**Prompt B: Query design.** *Using the vocabulary extracted above, draft search queries for each bibliographic source. List the component problems first, then provide at least one dedicated query for each problem and one query for each benchmark name. Avoid relying on broad umbrella terms when more specific terminology is available.*

**Prompt C: Relevance classification.** *Classify each record as highly relevant, somewhat relevant, or not relevant to the stated research questions. Treat a record as highly relevant only when its primary contribution addresses one of those questions. Identify the abstract sentence supporting the classification and provide a brief reason, including for negative classifications.*

**Prompt D: Extraction from prescriptive sources.** *Using only the supplied paper, record the problem addressed, model form, supervision, label assumptions, inspectable quantities, and limitations stated by the researchers. Identify the source location for each field. If a field is not addressed, mark it as not stated. Do not complete it from related work or from a possible extension of the method.*

**Prompt E: Extraction from descriptive sources.** *Using only the supplied source, record the finding or provision, the population to which it applies, the requirement it may impose on an automated system in this setting, and the evidence or authority on which it rests. Do not extrapolate a finding into a recommendation about model architecture.*

**Prompt F: Reconstruction of method families.** *Group the information extracted from each prescriptive paper by model form. For each family, state what it can and cannot express under two conditions: supervision is limited to one binary label for each item, and explanations should consist of quantities involved in the prediction. Report the relevant limitation and its supporting papers without recommending a family.*

**Prompt G: Gap and challenge review.** *Compare each draft gap claim with the supplied literature and identify evidence that narrows or weakens it. Then examine whether the proposed technical challenges follow from the verified gaps and domain requirements. Present alternative mappings and possible inconsistencies for author review. Do not make the final scope decision.*

Table A1 summarizes the main error categories covered by the literature checks. It describes the verification procedure and the action taken when an issue was identified. The table is not intended to cover every possible retrieval or writing error.

**Table A1. Literature Verification Errors and Safeguards**

| Error category | Typical source | Verification procedure | Disposition |
|---|---|---|---|
| Citation without a verifiable referent | Text generated by the model or incomplete notes | Match the citation to a bibliographic record before inclusion | Exclude until a verifiable source is located |
| Incorrect author, year, or venue | Metadata reconstructed from memory | Check against the bibliographic record and source document | Correct against the source record |
| Incorrect claim about publication status | Venue standing inferred from document format | Use bibliographic or venue metadata | Correct the status or cite the available preprint |
| Real source attached to an unsupported claim | Similarity between titles used as evidence | Read the relevant source passage and verify the specific claim | Remove the citation or replace the source |
| Finding applied outside its scope | Population, modality, or supervision conditions omitted | Review the source scope during structured extraction | Exclude the inference or state the relevant condition |
| Method capability inferred without supporting evidence | Potential extension treated as a reported result | Compare extracted fields with the paper's stated evaluation | Mark as not stated or narrow the characterization |

Most corrections occurred when retrieved records were converted into summaries or manuscript claims. We therefore repeated verification when a source was selected, when information was extracted, and when the citation was inserted into the manuscript. This process reduced reliance on a single initial check.

**A.2 Phase 2: AI-Augmented Data & Resource Construction**

Phase 2 translated the technical challenges established in Phase 1 into a validated resource base for subsequent design search. AI helped explore candidate representations, measurements, concepts, extractors, prompt formulations, and generic-safety resources, while the researchers assessed their domain validity, stability, specificity, and admissibility. These decisions determined what evidence and computational capabilities were available to Phase 3. Because AI4CDS is reversible, resource choices could later be revisited when design or evaluation evidence exposed weaknesses.

The final fixed measurement stage produces a frozen reading of general safety and a set of normalized concept measurements. The general safety reading is obtained from the frozen LlavaGuard judge and serves as the input to a trainable calibration. Semantic concepts are scored from the frozen visual representation with fixed prompt sets, while deterministic concepts are obtained from fixed extractors. Concepts used for contextual attenuation operate only on the general safety channel. Concepts related to risks for children are assigned in advance to the three mechanism blocks described in Appendix B.

The final assignments are M1 for imitable dangerous action, M2 for adultized performance, and M3 for self harm or depressive risk. M1 combines deterministic measurements of motion and pose with semantic evidence about risky action. M2 combines skin exposure, evidence of suggestive context, and five concepts related to adultized performance. M3 combines darkness, recall of self harm, and depressive affect. The activation structure is more selective than the full set of intensity measurements. M1 is activated by a family of motion measurements. M2 uses a conjunctive soft gate that combines skin exposure with the context of adultized performance. M3 is activated by darkness. Other assigned concepts affect the intensity of a mechanism but do not open its gate independently.

AI helped propose candidate concepts, prompt wording, extractor families, structural assignments, and diagnostics for weak or impure signals. The researchers assessed whether a proposed concept matched the developmental interpretation of its mechanism and whether its measurement was sufficiently stable and specific. They also determined whether the concept should affect activation, intensity, contextual attenuation, or none of these. Assignments to mechanism blocks and activation families were fixed before the risk assessment parameters were fitted.

Resource construction followed four checks.

1. **Construct validity**. The researchers assessed whether a proposed concept or measurement represented the developmental risk mechanism to which it would be assigned. Predictive association provided supporting evidence but did not establish construct validity on its own.
2. **Measurement behavior**. Candidate semantic scores and deterministic extractors were inspected for coverage, direction, missingness, and obvious contamination. When an instrument was blind to an important case, the measurement was revised or rejected. The downstream mechanism retained its original interpretation.
3. **Structural role**. Each retained signal was assigned to one role: the general safety reading, contextual attenuation, mechanism activation, or mechanism intensity. The final assignments followed Appendix B. Activation families may combine deterministic measurements with selected semantic context, while other semantic concepts contribute only to intensity.

4. **Separation of data partitions**. Normalization statistics and fixed measurement thresholds were estimated from the relevant training partition and then held fixed for data outside that partition. For the final fit, these quantities were estimated from the complete development set.

Resource diagnostics also constrained and sometimes reopened subsequent Phase 3 design choices. A proposed mechanism was removed when the available data did not support a distinct and estimable contribution. A mechanism for dangerous objects was rejected because the candidate measurements lacked sufficient specificity and stricter filtering left the signal too sparse. Short duration and stillness were excluded from the final deterministic inventory. Hard gate proposals were replaced by differentiable family gates. The design also distinguished evidence that opens a gate from evidence that affects only intensity. These findings led to corresponding changes in the architecture.

This phase also clarified the two forms of faithfulness addressed by the final design. Control over the input is supported by fixed concept definitions, structural assignments, caps on semantic concepts, and a semantic budget for each mechanism. Faithfulness in the mapping from evidence to prediction is supported by the anchored, monotone, and additive hazard construction. Appendix B reports the corresponding technical details, including normalization, gate structure, semantic projection, and the quantities used for explanation.

**A.3 Phase 3: AI-Generative Research Design Search & Refinement**

Phase 3 treated the three technical challenges established in Phase 1 as design requirements and searched for artifact designs that could realize them using the validated resources established in Phase 2. The artifact needed to preserve useful information about general safety while identifying additional risks to children. It also needed to represent different developmental risk mechanisms and maintain faithfulness in both the evidence entering the model and the mapping from evidence to prediction.

Before generating candidates, the researchers defined the design space and its main constraints. Reported contributions to risk for children should participate in the predictive computation. Mechanism labels should be grounded in developmental constructs. Unconstrained latent clusters did not meet this requirement. The method should operate without annotations at the concept level, and the original labels should remain unchanged. Normalization, model selection, and threshold selection should use development data only. Each retained design element should also be paired with an evaluation that could isolate its contribution. These conditions defined the admissible design space and were kept separate from predictive performance.

AI broadened the candidate set by proposing alternative model families, decompositions, response functions, constraints, and diagnostic tests. The researchers assessed whether each candidate addressed a stated technical challenge, fit the available supervision and resources, and had an independent justification. A candidate could be accepted, modified, reclassified, or rejected. When measurement or evaluation later revealed a problem, the relevant design decision was reconsidered. This review could lead to revision or removal of the component.

The design process involved five recurring operations.

1. **Translate each challenge into a testable design requirement**. The researchers specified the required property and the evidence that would count against it.
2. **Generate candidate mechanisms one component at a time**. AI proposed alternatives for the focal requirement. This kept each iteration focused and made the implications of each proposal easier to assess.
3. **Specify a corresponding evaluation before selection**. A candidate was retained when the researchers could define an ablation, diagnostic test, or theoretical condition that made its contribution assessable.
4. **Review claims and constructions critically**. Critiques generated with AI support helped identify unsupported guarantees, incompatible requirements, and arguments that no longer matched the current architecture. The researchers then decided whether to narrow the claim, revise the construction, or reject the component.
5. **Document consequential design decisions**. The record identified the origin of each candidate, the decision taken, and the basis for that decision. This information showed how the final architecture developed and helped prevent rejected alternatives from being reconsidered without new evidence.

Table A2 summarizes the main decisions that led to the final architecture specified in Appendix B. It distinguishes theoretical judgments from decisions based on measurement, available resources, or evidence from the development data.

**Table A2. Candidate Design Decisions and Dispositions**

| Candidate design | Origin | Disposition | Basis and consequence |
|---|---|---|---|
| Teacher distillation with counterfactual verification filtering | Authors | Rejected | The proposal primarily changed supervision construction and did not itself separate general safety from additional risk to children or make the explanation part of the prediction. |
| Separate adult and child judgments for contrastive supervision | Authors | Rejected | With the available binary labels, the contrast did not provide a distinct learning target for the intended mechanism. |
| Mixture of experts as the base formulation | Authors | Reclassified | Retained as an architectural structure for components associated with specific mechanisms. It did not serve as the methodological foundation. |
| Concept bottleneck family without concept labels | Authors | Accepted as the starting family | Among the reviewed families, it provided the closest fit to supervision at the item level and prediction through concepts without requiring concept labels. |
| Direct classifier for risk to children without residual decomposition | AI | Rejected | It did not preserve a distinct and reusable channel for general safety or define the additional risk to children relative to that channel. |
| Signed residuals in logit space | AI | Rejected | Signed contributions allowed cancellation and weakened the intended nonnegative decomposition. The design moved to additive nonnegative hazards. |
| Fixed transformation of the frozen general safety reading | AI | Modified | Diagnostics on the development data motivated a bounded and learned monotone calibration, together with a separate mechanism that attenuates the general channel according to context. |
| Auxiliary evidence about general safety used in the separation objective | Authors and AI | Accepted after refinement | Frozen auxiliary evidence was retained for the separation objective during training. It was not treated as the channel that predicts general safety. |
| Seven developmental vulnerability mechanisms | Authors | Reduced | Available data and measurement support did not sustain that level of differentiation. The final specification uses M1, M2, and M3. |
| Experts partitioned by signal type | AI | Rejected | A partition based on modality did not provide the developmental interpretation required of the mechanism labels. |
| Four mechanism experts | Authors and AI | Reduced | Repeated development analysis did not support one proposed mechanism, leaving the three mechanisms specified in Appendix B. |
| Fixed hard gates | Authors and AI | Modified | Replaced by continuous differentiable gates with learned family thresholds. The same soft gate is used during training and inference. |
| Expert for dangerous objects | AI | Rejected after measurement review | Candidate measurements were not sufficiently specific, and stricter filtering left too little usable evidence for a separate mechanism. |
| Interaction terms between mechanism hazards | AI | Removed after diagnosis | Simultaneous activation was too sparse to support stable interaction terms. The final hazard decomposition therefore remains additive. |
| Unconstrained contributions from semantic concepts | Default of the base family | Rejected | Unbounded semantic evidence at the measurement stage would weaken faithfulness on the input side. The final design uses a cap for each concept, a semantic budget, and projection during inference. |
| A single aggregate cap on semantic evidence | AI | Modified | Refined into the dominance and allowance constraint at the mechanism level and the concept caps described in Appendix B. |
| Unconstrained expert response functions | Default of the base family | Rejected | Development diagnostics showed that unconstrained responses could conflict with the direction implied by a mechanism's interpretation. |
| Monotone and bounded concept responses | AI | Accepted | The anchored monotone response basis preserves the declared direction of risk while allowing nonlinear growth, threshold behavior, and saturation. |
| More flexible nonadditive expert functions | AI | Rejected | Additional flexibility would weaken direct reconstruction of contributions from mechanisms and concepts. |
| Shared intercept absorbed within concept responses | AI | Modified | The final model separates baseline risk that does not depend on the input from responses to specific concepts through a nonnegative baseline hazard. |
| Abstention whenever no mechanism gate is active | AI | Modified | The initial rule withheld too many decisions. The final artifact produces a prediction from the complete computation of general and child risk. |
| Direct reuse of an existing identifiability result | Authors | Narrowed and rederived | The final claims were restricted to the additive model class in Proposition 2, in which blocks are separable and responses are anchored at zero. |
| Separate novelty claims for concept identity and attribution | Authors and AI | Consolidated | The claims were restated as a graded account of control over the input and faithfulness in the mapping from evidence to prediction. |

These decisions suggest that AI contributed more to generating candidates than to selecting the final design. Many candidates were revised or rejected after theoretical review, measurement analysis, or diagnostics on the development data. This pattern is consistent with the intended division of labor in AI4CDS. AI can broaden the design search, while the researchers remain responsible for admissibility and scientific judgment.

**A.4 Phase 4: AI-Enabled Multi-Faceted Evaluation**

Phase 4 implemented and evaluated the selected artifact. AI helped draft execution specifications, implement the proposed model, generate diagnostic analyses, and suggest possible interpretations. The researchers set the evaluation protocol, reviewed the specifications before execution, checked the outputs before interpretation, and verified that the implementation corresponded to Appendix B.

Normalization, parameter fitting, hyperparameter selection, and threshold selection used only development data. The operating threshold was selected from predictions generated through fivefold cross validation within the development set. A final risk assessment model was then fitted on the complete development set. Its learned parameters and threshold were fixed before evaluation on the test set. Metrics at the selected operating point and complementary ranking metrics were reported as distinct views of performance.

Specification, execution, and interpretation were treated as separate steps. An execution specification stated the objective, the intended change, the quantities that should remain fixed, and the outputs to be retained. The researchers reviewed the specification before it was carried out. They then checked the reported values and artifacts before accepting an interpretation. This separation made differences among the designed method, the implemented code, and the reported claims easier to identify.

At the two main handovers, the researchers reviewed the proposal and reached agreement before the process continued. Before execution, they discussed the research question, permitted change, evaluation rule, and evidence to be retained. After execution, they reviewed the outputs together and provided feedback on missing checks, conflicting results, and claims that exceeded the evidence. AI then supported additional diagnostics or revised analyses when needed. The researchers retained interpretations that they considered consistent with the design specification and the reported evidence.

The implementation record was organized around five checks.

1. **Define the evaluation rules in advance**. Data partitions, primary metrics, threshold selection, and comparison rules were set without using test labels for model selection.
2. **Change one focal quantity at a time**. Ablations and diagnostics were specified so that the effect of a component could be attributed to the intended change. Checks on quantities outside the authorized change were reported. Any further correction was handled in a separate step.
3. **Reconcile code and design in both directions**. Code was treated as evidence of what was executed, while Appendix B was treated as the authoritative statement of what was being claimed. When the two diverged, the implementation, description, or claim was revised explicitly.
4. **Evaluate benchmarks under comparable operating rules**. AI supported benchmark selection and checks of comparability. The researchers reviewed benchmark implementations and reported comparisons. Operating thresholds were selected with development data held out from the corresponding model fit. This procedure avoided comparisons based on different default thresholds.
5. **Keep interpretation under author responsibility**. Explanations of results generated with AI were treated as proposals. The researchers examined conflicting evidence across splits, unsupported causal interpretations, and apparent anomalies. Their feedback could lead to further diagnostics, a narrower claim, or rejection of the proposed interpretation.

Table A3 summarizes selected issues in implementation or interpretation that led to a correction, rejection, or qualification. It does not cover every possible implementation risk.

**Table A3. Evaluation Issues and Researcher Dispositions**

| Item | Origin | Issue identified | Disposition |
|---|---|---|---|
| Analysis rows based on an operating point informed by test data | Analysis proposed by AI | The operating point relied on information reserved for final evaluation | Rows removed; operating points restricted to development data |
| Abstention when no gate for child risk was active | Component proposed by AI | The rule withheld decisions for a substantial part of the evaluation set | Rule revised; the final artifact returns a prediction from the complete hazard computation |
| Proposal to alter ambiguous labels to purify a signal | Test of the safeguard | The action conflicted with the rules for preserving labels and research integrity | Not executed; labels remained unchanged |
| Gate described as a pure disjunction | Manuscript text drafted with AI support | The description did not match the conjunctive family structure used by the M2 soft gate | Text corrected against the implementation and Appendix B |
| Architecture for general safety described inconsistently across materials | Reporting artifact produced by an author | The description obscured the frozen general safety reading and the separate evidence used only during training for residual separation | Reporting reconciled with Appendix B before interpretation |
| Component contribution reported without a corresponding qualification about uncertainty | Analysis supported by AI | The evidentiary standard differed from that applied to another quantity computed on a similarly small activated subset | The claim was qualified and the same standard applied across the compared quantities |
| Recommendation to remove a mechanism based on one split | Interpretation generated by AI | Evidence from the development data did not consistently support the recommendation | Recommendation not adopted; disagreement retained as diagnostic evidence |
| Very little variation across repeated fits treated as a reporting error | Initial reading of a diagnostic result | The behavior was consistent with a small risk assessment layer fitted over frozen measurements | Explained and checked; no automatic correction was applied |

Two patterns informed the framework. First, important corrections were often identified when work moved from specification to execution or from execution to interpretation. Second, several safeguards concerned analyses that could have made the results appear stronger, such as operating points informed by test data, broad abstention, or point estimates reported without qualification. These observations suggest that human review is particularly important when a computational output becomes an experimental result or a scientific claim.

The evaluation reported in Section 4 follows the protocol summarized above. Predictive comparisons assess artifact utility at the selected operating point. Ablations examine the contribution of the four design elements, and changes in the representation assess robustness. The interpretation cases trace decisions to hazards related to general safety and risks to children, together with the realized concept contributions from the same forward computation. Appendix B specifies the final model and training procedure. This appendix describes how AI support and human judgment contributed to its design, implementation, and verification.

### A.5 Phase 5: AI-Assisted Design Principle Distillation

Phase 5 consolidated the research record accumulated across Phases 1–4 and used that record together with the evaluation evidence to distinguish case-specific findings from transferable design knowledge. Documentation was therefore not reconstructed only at the end of the project; the consequential decisions, alternatives, corrections, and verification records documented in the preceding phases provided the evidence base for Phase 5. AI supported provenance organization, cross-phase comparison, and identification of candidate patterns, while the researchers verified the record, assessed the scope of candidate abstractions, and determined which findings warranted claims beyond the ChildRiskGuard case.

The distillation process involved five explicit operations. A candidate abstraction could be revised, narrowed, rejected, or returned to an earlier operation when the supporting evidence was insufficient.

1. **Assemble the cross-phase research record**. The records from Phases 1–4 were consolidated around consequential research transitions, including the domain requirements and technical challenges, admitted and rejected resources, retained and rejected design alternatives, evaluation evidence, corrections, and researcher interventions. AI helped organize these materials and connect related records across phases. The researchers checked that consequential decisions were represented accurately and remained traceable to the underlying evidence.
2. **Link consequential decisions to evidence and outcomes**. The researchers examined how requirements established in Phase 1 constrained resources in Phase 2 and candidate designs in Phase 3, and how Phase 4 evidence supported, revised, qualified, or rejected earlier decisions. AI supported comparison across these records and identification of possible decision–evidence links. The researchers verified the links before they were used in subsequent abstraction.
3. **Surface candidate cross-phase patterns**. AI helped identify recurring relationships among domain requirements, resource choices, design decisions, evaluation results, corrections, and researcher interventions. It also proposed candidate formulations of methodological or design insights that could summarize these relationships. These candidate patterns were treated as proposals rather than as established findings.

4. **Evaluate candidate abstractions and define their scope**. The researchers assessed each candidate abstraction against the documented research record and the available evaluation evidence. They examined whether the proposed relationship depended on features specific to ChildRiskGuard, whether contradictory cases were present, and whether the evidence supported the level of generality implied by the formulation. Unsupported abstractions were rejected, while partially supported abstractions were narrowed or accompanied by explicit boundary conditions. AI supported critical comparison and alternative formulations but did not determine which claims were retained.
5. **Distill and document transferable design knowledge**. The researchers assessed which surviving patterns warranted abstraction as transferable design insights, methodological principles, governance principles, reproducibility guidance, or boundary conditions, using established theory to explain the underlying mechanisms and discipline the scope of generalization. AI supported organization, comparison, and refinement of candidate formulations, while the researchers retained responsibility for their substantive interpretation, level of abstraction, and final wording. The resulting claims were linked both to their theoretical grounding and to the cross-phase evidence that demonstrated their operation in ChildRiskGuard.

## Appendix B Detailed Method

### B.1 Measurement and Concept Specification

#### B.1.1 Frozen Video and Generic-Safety Measurement

For video $V_i$, uniformly sample $T$ frames,

$$X_i = \{x_{i,1}, ..., x_{i,T}\}. \tag{B1}$$

Let $\phi_{LG}$ denote the frozen LlavaGuard vision tower. The supplied workflow reports that this tower was verified to match the OpenAI Contrastive Language–Image Pre-training (CLIP) ViT-L/14-336 model after casting the original 32-bit floating-point weights to the 16-bit floating-point precision stored in LlavaGuard; all 391 vision-tower tensors matched. For frame $x_{i,t}$,

$$e_{i,t} = \phi_{LG}\left(x_{i,t}\right). \quad \text{(B2)}$$

The frozen LlavaGuard judge returns sequence log-probabilities for the Safe and Unsafe responses,

$$J_{LG}\left(x_{i,t}\right) \rightarrow \left(\ell_{i,t}^{safe}, \ell_{i,t}^{unsafe}\right). \quad \text{(B3)}$$

The corresponding frame-level unsafe probability is

$$q_{i,t} = \frac{exp\left(\ell_{i,t}^{unsafe}\right)}{exp\left(\ell_{i,t}^{safe}\right)+exp\left(\ell_{i,t}^{unsafe}\right)}. \quad \text{(B4)}$$

We aggregate across sampled frames by

$$q_i = \max_t q_{i,t}, \qquad u_i = - log\left(1 - q_i\right). \quad \text{(B5)}$$

Thus, $u_i$ is the frozen generic-safety reading used by the trainable risk-assessment stage; it is not the child-inappropriateness probability.

B.1.2 Semantic Concept Measurement

For semantic concept $k$, let $P_k^+$ and $P_k^-$ denote its fixed positive and negative prompt sets. For prompt $p$, the frozen CLIP text encoder produces the unit-normalized representation

$$r_p = Norm_2\left(\psi_{CLIP}(p)\right). \quad \text{(B6)}$$

Each frame embedding is first unit-normalized. The video representation is the renormalized temporal mean

$$\bar{e}_i = Norm_2\left(\frac{1}{T}\sum_{t=1}^{T} Norm_2\left(e_{i,t}\right)\right). \quad \text{(B7)}$$

The broad risky-action, self-harm, and contextual-attenuation concepts use the temperature-scaled positive-versus-negative score

$$c_{ik}^{sem} = \sum_{p\in P_k^+} \frac{exp\left(\tau_{CLIP}\bar{e}_i^\top r_p\right)}{\sum_{q\in P_k^+\cup P_k^-} exp\left(\tau_{CLIP}\bar{e}_i^\top r_q\right)}, \qquad \tau_{CLIP} = 20. \quad \text{(B8)}$$

The five adultized-performance concepts use the positive-prompt maximum cosine score

$$c_{ik}^{sem} = \max_{p\in P_k^+} \bar{e}_i^\top r_p. \quad \text{(B9)}$$

Their aggregate context measurement is

$$c_i^{adultized} = \max_{k\in K_{adultized}} c_{ik}^{sem}, \quad \text{(B10)}$$

where $K_{adultized}$ contains sexualized dance, nightclub dance, revealing outfit, suggestive pose, and adult styling. Thus, LlavaGuard judge scores use temporal maximum pooling, while semantic concepts use one normalized mean video embedding followed by a frozen prompt-set scorer.

B.1.3 Deterministic Concept Measurement

A deterministic concept is obtained from a fixed reproducible extractor,

$$c_{ik}^{det} = D_k(V_i). \quad \text{(B11)}$$

The final deterministic inventory contains body motion, global flow, flow peak, motion energy, and pose inversion for M1; skin exposure for M2; and darkness for M3. Short duration and stillness are excluded from the final model.

B.1.4 Normalization

For each fit, let $\ell_k$, $h_k$, and $m_k$ be the 0.05 quantile, 0.95 quantile, and median of concept $k$ on that fit's training partition. Define the oriented endpoints as $a_k = \ell_k$ and $b_k = h_k$ for an increasing-risk concept, and $a_k = h_k$ and $b_k = \ell_k$ for a decreasing-risk concept. Missing or non-finite values are replaced by $m_k$. Every concept is then normalized to $[0, 1]$ by

$$\tilde{c}_{ik} = clip\left(\frac{c_{ik} - a_k}{b_k - a_k}, 0, 1\right). \quad \text{(B12)}$$

If the two quantiles coincide, the observed minimum and maximum are used with the same orientation; a unit-width fallback is used only for a constant feature. The fitted normalization map is frozen for held-out and test data. In the final full-development fit, the complete development set is the training partition.

B.1.5 Concept Inventory and Fixed Structural Assignments

The final mechanism assignments are as follows. M1 uses the five deterministic motion/pose concepts above and a risky-action semantic concept. M2 uses skin exposure, a broad suggestive-context concept, and five adultized-performance concepts: sexualized dance, nightclub dance, revealing outfit, suggestive pose, and adult styling. M3 uses darkness, self-harm recall, and depressive affect. The contextual attenuation set $E$ contains animation/cartoon, game interface, news report, educational/medical, and documentary concepts; these operate only on the generic-safety channel.

The manuscript assumes

$$C_r \cap C_s = \varnothing \quad (r \neq s), \qquad E \cap \left(\cup_{r=1}^{R} C_r\right) = \varnothing. \tag{B13}$$

## B.2 Model Architecture and Training

### B.2.1 Anchored Monotone Responses and Generic-Safety Channel

*Anchored monotone response basis.*

After robust normalization to $[0, 1]$, all concepts use the same evenly spaced knot locations $\kappa_q = q/Q$, $q = 0, ..., Q$. Define

$$b_q(x) = min\left\{\left(x - \kappa_{q-1}\right)_+, \kappa_q - \kappa_{q-1}\right\}. \tag{B14}$$

The concept response used in mechanism $r$ is

$$h_{rk}(x) = \sum_{q=1}^{Q} \omega_{rkq} b_q(x), \qquad \omega_{rkq} \geq 0. \tag{B15}$$

The implementation parameterizes the learned knot-value increments through a positive transform and scales them by a learned bounded cap. This is equivalent to nonnegative segment slopes, so $h_{rk}(0) = 0$ and $h_{rk}$ is nondecreasing and bounded. Different segment slopes permit nonlinear threshold-like growth and saturation while retaining zero anchoring and monotonicity.

The final implementation uses $Q = 8$ knots and a maximum per-concept response of 8.0 before the tighter semantic cap is applied. Positive increments are implemented with a softplus transform and normalized cumulative sums; a sigmoid-parameterized cap bounds each terminal knot value.

*Generic-safety calibration.*

Let $u_{ref}$ and $u_{high}$ be the $0.05$ and $0.95$ training-distribution quantiles of the frozen generic reading. The input to the generic calibrator is

$$u_i^{norm} = clip\left(\frac{u_i - u_{ref}}{u_{high} - u_{ref}}, 0, 1\right). \tag{B16}$$

The calibrator $\psi_G$ linearly interpolates eight learned knot values at evenly spaced locations on $[0, 1]$. Its knot increments are positive, its first knot is zero, and its learned terminal value is bounded by the generic-hazard cap $H_G = 4.5$. Hence $\psi_G(0) = 0$ and $0 \le \psi_G\left(u_i^{norm}\right) \le 4.5$. In the final full-development fit, $u_{ref} = 0.02621335$ and $u_{high} = 0.19349697$.

*Contextual attenuation.*

Let $E$ be the fixed set of contextual attenuation concepts. The contextual attenuation factor is

$$a_i = \prod_{e \in E} \sigma\left(\frac{\vartheta_e - \tilde{c}_{ie}}{\tau_e}\right), \tag{B17}$$

and the generic-safety hazard is

$$\lambda_i^G = a_i \psi_G\left(u_i^{norm}\right). \tag{B18}$$

The contextual attenuation concepts are frozen readouts from the same visual representation used for semantic concept scoring and do not enter any child-specific mechanism block. Each $\vartheta_e$ is fixed at the $0.90$ training-distribution quantile of concept $e$, and the final model fixes $\tau_e = 0.25$ for all contextual concepts. Neither quantity is learned by gradient descent.

B.2.2 Mechanism Structuring and Differentiable Gating

For mechanism $r$, the knowledge-anchored concept block is

$$C_r = C_r^{det} \cup C_r^{sem}, \tag{B19}$$

and its activation evidence is divided into fixed families $\{A_{rf}\}_{f=1}^{F_r}$ with $A_{rf} \subseteq C_r$. Both concept-block and family memberships are fixed before training.

Within family $f$, the smooth anchor summary is

$$a_{irf} = \frac{1}{\beta_g} log\left(\sum_{k \in A_{rf}} exp\left(\beta_g \tilde{c}_{ik}\right)\right), \tag{B20}$$

and the family response and mechanism activation are

$$z_{irf} = \sigma\left(\frac{a_{irf} - \vartheta_{rf}}{\tau_g}\right),\ \gamma_i^r = \prod_{f=1}^{F_r} z_{irf}. \tag{B21}$$

The smooth maximum is coordinatewise nondecreasing within each family, and the cross-family product is coordinatewise nondecreasing because

$$\frac{\partial \gamma_i^r}{\partial z_{irf}} = \prod_{g \neq f} z_{irg} \geq 0. \tag{B22}$$

The final gate topology is: one M1 family containing body motion, global flow, flow peak, and motion energy; two M2 families containing skin exposure and the five adultized-performance concepts, respectively; and one M3 family containing darkness. Families are combined conjunctively, so M2 implements a soft skin-exposure–adultized-context gate. Pose inversion, broad suggestive recall, self-harm recall, and depressive affect are intensity evidence rather than independent gate-opening evidence.

The mechanism intensity and mechanism hazard are

$$\mu_i^r = \mu_i^{r,det} + \mu_i^{r,sem}, \qquad \lambda_i^r = \gamma_i^r \mu_i^r. \tag{B23}$$

The same continuous gate is used during training and inference. The gate temperature follows a log-linear schedule from 0.30 toward 0.05 over the first 80% of the configured training horizon; the fixed epoch-480 checkpoint has $\tau_g = 0.10261319$. The within-family smooth-maximum parameter is $\beta_g = 8$.

B.2.3 Semantic Contribution Budget

The mechanism intensity is partitioned into

$$\mu_i^{r,det} = \sum_{k \in C_r^{det}} h_{rk}\left(\tilde{c}_{ik}\right), \ \mu_i^{r,sem} = \alpha_{sem} \sum_{k \in C_r^{sem}} min\{h_{rk}\left(\tilde{c}_{ik}\right), H_{sem}\}, \tag{B24}$$

with $\mu_i^r = \mu_i^{r,det} + \mu_i^{r,sem}$.

The final model uses $\alpha_{sem} = 0.35$ and $H_{sem} = 2.5$. Its semantic contribution constraint is

$$\mu_i^{r,sem} \leq \kappa_r^{dom} \mu_i^{r,det} + s_r. \tag{B25}$$

The corresponding share parameterization is

$$\beta_r = \frac{\kappa_r^{dom}}{1+\kappa_r^{dom}}, \qquad \delta_r = \frac{s_r}{1+\kappa_r^{dom}}. \tag{B26}$$

The final configuration fixes $\kappa_r^{dom} = 1$ for all mechanisms and $\left(s_1, s_2, s_3\right) = (1.0, 1.0, 1.2)$.

The pointwise violation is

$$b_i^r = \left[\mu_i^{r,sem} - \kappa_r^{dom}\mu_i^{r,det} - s_r\right]_+, \tag{B27}$$

and the implementation uses the mean unsquared-hinge training penalty

$$L_{bud} = \frac{1}{D}\sum_{i=1}^{D}\sum_{r=1}^{R} b_i^r. \tag{B28}$$

The primary model additionally enforces the same constraint by hard projection at inference:

$$\mu_i^{r,sem} \leftarrow min\left\{\mu_i^{r,sem}, \kappa_r^{dom}\mu_i^{r,det} + s_r\right\}. \tag{B29}$$

The per-semantic-concept cap used in the primary model is

$$h_{rk}^{cap}(x) = min\{h_{rk}(x), H_{sem}\}, \tag{B30}$$

with $H_{sem} = 2.5$. It preserves monotonicity while bounding each individual semantic response. Thus the per-concept cap, soft budget penalty, and inference projection are all active in the final model.

B.2.4 Joint Training, Calibration, and Faithful Output

*Trainable and frozen components.*

The measurement stage is frozen:

$$\nabla J_{LG} = 0,\ \nabla\phi_{LG} = 0,\ \nabla\psi_{CLIP} = 0,\ \nabla D_k = 0. \tag{B31}$$

Only the risk-assessment stage is optimized. Its trainable parameters include the nonnegative response coefficients, generic-safety calibration coefficients, mechanism-family gate thresholds, and the nonnegative leak hazard $\lambda^0$. The contextual attenuation thresholds are fixed training-distribution quantiles rather than trainable parameters.

The dominance ratios $\kappa_r^{dom}$, slacks $s_r$, semantic scale, semantic cap, generic-hazard cap, loss weights, and gate-temperature schedule are fixed hyperparameters in the final fit. The mechanism-family thresholds are learned. The final probability uses identity hazard temperature $T = 1$, and the operating threshold is selected from five-fold out-of-fold development predictions rather than from the test set.

*Training objective.*

For probability $p_i = 1 - exp\left(-\Lambda_i\right)$, the implementation uses class-weighted binary cross-entropy

$$L_{BCE} = -\frac{1}{D}\sum_{i=1}^{D}\left[w_+ y_i log p_i + \left(1 - y_i\right)log\left(1 - p_i\right)\right], \qquad w_+ = 0.45\frac{D_-}{D_+}, \tag{B32}$$

where $D_+$ and $D_-$ are the positive and negative counts in the training partition. The realized-hazard sparsity term is

$$L_{sparse} = \frac{1}{DR}\sum_{i=1}^{D}\sum_{r=1}^{R}\lambda_i^r. \quad \text{(B33)}$$

It discourages unnecessarily large or diffuse child-specific mechanism hazards; it is not a discrete concept-selection penalty.

The unified objective, using the same loss-weight notation as the main text, is

$$L = L_{BCE} + \xi_{sp}L_{sparse} + \xi_{bud}L_{bud} + \xi_{sep}L_{sep}. \quad \text{(B34)}$$

*Residual-separation penalty.*

Let $s_i$ be the four-dimensional frozen generic-safety evidence vector containing calibrated visual-sexual, visual-violence, automatic speech recognition (ASR)-toxicity, and optical character recognition (OCR)-toxicity readings. For minibatch $B$, define

$$A_{r,B} = \{i \in B: \gamma_i^r \geq 0.5\}. \quad \text{(B35)}$$

The implemented separation term is

$$L_{sep} = \frac{1}{|R_B|}\sum_{r\in R_B} nHSIC\left(\lambda^r_{A_{r,B}}, S_{A_{r,B}}\right), \quad \text{(B36)}$$

where $R_B$ contains mechanisms with at least 20 active samples in the minibatch. Let $K$ and $L$ be the radial-basis-function Gram matrices for the active mechanism hazards and generic-safety vectors. Each RBF bandwidth is the median positive pairwise squared distance. With $H = I - m^{-1}11^{\top}$ for $m$ active samples,

$$HSIC(K, L) = \frac{1}{(m-1)^2}tr(KHLH), \quad \text{(B37)}$$

and

$$nHSIC(K, L) = \frac{HSIC(K,L)}{\sqrt{max\{HSIC(K,K)HSIC(L,L),\varepsilon\}}}. \quad \text{(B38)}$$

This is a soft dependence regularizer rather than an independence guarantee.

*Probability calibration and decision rule.*

The final model fixes the hazard temperature to identity and computes

$$\hat{p}_i = 1 - exp\left(-\Lambda_i\right). \quad \text{(B39)}$$

The moderation decision is

$$\hat{y}_i = I\left[\hat{p}_i \geq \eta\right], \quad \text{(B40)}$$

where $\eta$ is selected by maximizing the F1 score on five-fold out-of-fold predictions from the combined development set. With fixed epoch 480, the selected threshold is $\eta = 0.6123511$. A single model is then fitted on the complete development set, and the locked threshold is applied once to the test set.

*Faithful hazard output.*

For $\Lambda_i > 0$, the mathematically defined hazard shares are

$$\pi_i^G = \frac{\lambda_i^G}{\Lambda_i}, \qquad \pi_i^r = \frac{\lambda_i^r}{\Lambda_i}, \qquad \pi_i^0 = \frac{\lambda^0}{\Lambda_i}. \quad \text{(B41)}$$

If $\Lambda_i = 0$, all shares are defined as zero. In software, a small numerical stabilizer may be used only to avoid division by zero; it should not replace the mathematical definition above when discussing exact calibration invariance.

At concept level,

$$\phi_{ik}^r = \{\gamma_i^r h_{rk}\left(\tilde{c}_{ik}\right),\ k \in C_r^{det},\ \gamma_i^r \chi_i^r \alpha_{sem} min\{h_{rk}\left(\tilde{c}_{ik}\right), H_{sem}\},\ k \in C_r^{sem}, \qquad \lambda_i^r = \sum_{k \in C_r} \phi_{ik}^r. \quad \text{(B42)}$$

**B.3 Theoretical Results and Proofs**

B.3.1 Proof of Proposition 1

Let $F: R_+ \rightarrow [0, 1)$ satisfy the assumptions in Proposition 1 and define $Q(t) = 1 - F(t)$. Then $Q(0) = 1$, $Q(t) > 0$, and

$$Q(a + b) = 1 - F(a + b) = [1 - F(a)][1 - F(b)] = Q(a)Q(b). \quad \text{(B43)}$$

Because $F$ is continuous and strictly increasing, $Q$ is continuous and strictly decreasing. Define $G(t) =- logQ(t)$. Then

$$G(a + b) = G(a) + G(b). \quad \text{(B44)}$$

The continuous solutions of the additive Cauchy equation are $G(t) = ct$ for some constant $c$. Strict increase of $F$ implies $c > 0$. Hence

$$Q(t) = exp(-\, ct), \qquad F(t) = 1 - exp(-\, ct). \tag{B45}$$

Rescaling every additive risk quantity by $c$ gives the canonical form $F(t) = 1 - exp(-\, t)$.

B.3.2 Proof of Proposition 2

Write the total hazard as

$$\Lambda(x) = \lambda^0 + f_G\left(x_G\right) + \sum_{r=1}^{R} f_r\left(x_r\right), \tag{B46}$$

where $x_G = \left(u, c_E\right)$ contains the generic-safety reading and contextual attenuation concepts, and $x_r = c_{C_r}$ contains the concepts assigned to mechanism $r$. Let the designated reference points be $x_G^0$ and $x_r^0$, and assume

$$f_G\left(x_G^0\right) = 0, \qquad f_r\left(x_r^0\right) = 0. \tag{B47}$$

Suppose a second decomposition

$$\widetilde{\Lambda}(x) = \widetilde{\lambda}^0 + \widetilde{f}_G\left(x_G\right) + \sum_{r=1}^{R} \widetilde{f}_r\left(x_r\right) \tag{B48}$$

produces the same total hazard for every admissible input and satisfies the same block and anchoring restrictions.

Setting every block to its reference value gives $\lambda^0 = \widetilde{\lambda}^0$. Varying only the generic block while all mechanism blocks remain at reference gives $f_G\left(x_G\right) = \widetilde{f}_G\left(x_G\right)$. For any mechanism $r$, varying only $x_r$ while all other blocks remain at reference gives $f_r\left(x_r\right) = \widetilde{f}_r\left(x_r\right)$. The required blockwise variations exist by the joint-support assumption. Therefore all components coincide.

For ChildRiskGuard, $f_G$ is the context-adjusted generic-safety hazard, and each $f_r$ contains the soft mechanism gate and anchored mechanism intensity. At the all-zero mechanism reference, $\mu_i^r = 0$, hence $\lambda_i^r = \gamma_i^r \mu_i^r = 0$ regardless of the gate value. At the generic reference, $\psi_G\left(u_{ref}\right) = 0$, hence $\lambda_i^G = 0$ regardless of contextual attenuation.

B.3.3 Proof of Proposition 3

If the pointwise semantic contribution constraint holds, then

$$\mu_i^{r,sem} \leq \beta_r \mu_i^r + \delta_r. \quad \text{(B49)}$$

Multiplying by $\gamma_i^r \geq 0$ yields

$$\gamma_i^r \mu_i^{r,sem} \leq \beta_r \lambda_i^r + \gamma_i^r \delta_r. \quad \text{(B50)}$$

Summing over mechanisms gives

$$H_i^{sem} \leq \sum_r \beta_r \lambda_i^r + \sum_r \gamma_i^r \delta_r \leq \beta_{max} H_i^{child} + \sum_r \gamma_i^r \delta_r. \quad \text{(B51)}$$

If every $\delta_r = 0$ and $H_i^{child} > 0$, division by $H_i^{child}$ yields the semantic-share bound.

B.3.4 Formal Faithfulness Axioms and Proposition 4

The main text introduces Axiom 1–Axiom 4 (A1-A4) as the operational criteria for mapping-side explanatory faithfulness. Their formal definitions are as follows.

*A1 (Completeness).*

For every video $i$ and mechanism $r$,

$$\lambda_i^r = \sum_{k \in C_r} \phi_{ik}^r, \qquad \phi_{ik}^r = \{\gamma_i^r h_{rk}\left(\tilde{c}_{ik}\right),\ k \in C_r^{det},\ \gamma_i^r \chi_i^r \alpha_{sem} min\{h_{rk}\left(\tilde{c}_{ik}\right), H_{sem}\},\ k \in C_r^{sem}. \quad \text{(B52)}$$

*A2 (Sign consistency).*

For each child-risk concept $k \in C_r$, holding the other concept measurements fixed,

$$\frac{\partial \lambda_i^r}{\partial \tilde{c}_{ik}} \geq 0. \quad \text{(B53)}$$

Contextual attenuation concepts have the opposite declared direction, but are restricted to the generic-safety channel and are not child-risk concepts in $C_r$.

*A3 (Unique attribution).*

Every concept contribution has one structural destination: the child-specific mechanism blocks are mutually disjoint, the contextual attenuation block is separate from all mechanism blocks, and the additive channel decomposition is identifiable under Proposition 2.

*A4 (Non-masking).*

Let $\Lambda_i^{(-r)}$ be the total hazard with mechanism $r$ removed. Then

$$\Lambda_i - \Lambda_i^{(-r)} = \lambda_i^r \geq 0. \tag{B54}$$

*Proof of Proposition 4: Axiomatic faithfulness of the child-specific pathway.*

We verify the four axioms under the conditions stated in the main text.

A1. By the definitions of projected mechanism intensity and realized concept contribution,

$$\sum_{k \in C_r} \phi_{ik}^r = \gamma_i^r \left( \mu_i^{r,det} + \overline{\mu}_i^{r,sem} \right) = \gamma_i^r \mu_i^r = \lambda_i^r. \tag{B55}$$

A2. Before aggregate projection, every deterministic or capped-and-scaled semantic response is nondecreasing in its own normalized concept. For a concept $k$ outside the activation families, $\gamma_i^r$ is constant with respect to $\tilde{c}_{ik}$ and therefore

$$\frac{\partial \lambda_i^r}{\partial \tilde{c}_{ik}} \geq 0. \tag{B56}$$

For a concept that belongs to an activation family,

$$\frac{\partial \lambda_i^r}{\partial \tilde{c}_{ik}} = \frac{\partial \gamma_i^r}{\partial \tilde{c}_{ik}} \mu_i^r + \gamma_i^r \frac{\partial \mu_i^r}{\partial \tilde{c}_{ik}} \geq 0, \tag{B57}$$

because the smooth family summary, soft threshold, cross-family product, mechanism intensity, and concept response are all nonnegative/nondecreasing in the relevant argument. Under inference projection, the semantic aggregate is the minimum of two nondecreasing nonnegative quantities, so the mechanism hazard remains nondecreasing; at the projection kink, the statement is understood in the one-sided/subgradient sense.

A3. Equation (B13) assigns every child-risk concept to at most one mechanism block and separates contextual attenuation concepts from the child-specific blocks. Proposition 2 makes the resulting channel-level decomposition unique under the stated block-support and anchoring conditions. Thus a concept contribution cannot be arbitrarily reassigned to a different mechanism or the generic-safety channel.

A4. The total hazard is additive and every mechanism hazard is nonnegative. Removing mechanism $r$ therefore changes the total hazard by exactly

$$\Lambda_i - \Lambda_i^{(-r)} = \lambda_i^r \geq 0. \tag{B58}$$

Thus all four axioms hold for the child-specific pathway under the stated conditions.

B.3.5 Conditional Shapley Correspondence in the Additive Regime

When the aggregate semantic projection is inactive, $\chi_i^r = 1$, the additive mechanism-intensity pathway has an exact game-theoretic interpretation conditional on the realized mechanism gate. Fix $\gamma_i^r$ and define, for $S \subseteq C_r$,

$$v_i^r(S) = \gamma_i^r \sum_{k \in S} \tilde{h}_{rk}\left(\tilde{c}_{ik}\right), \tag{B59}$$

where $\tilde{h}_{rk} = h_{rk}$ for deterministic concepts and $\tilde{h}_{rk} = \alpha_{sem} min\{h_{rk}, H_{sem}\}$ for semantic concepts. For any coalition $S$ not containing $k$,

$$v_i^r(S \cup \{k\}) - v_i^r(S) = \gamma_i^r \tilde{h}_{rk}\left(\tilde{c}_{ik}\right) = \phi_{ik}^r. \tag{B60}$$

The marginal contribution of concept $k$ is therefore identical for every coalition not containing it. Its Shapley value is consequently

$$Shap_{ik}^r = \phi_{ik}^r. \tag{B61}$$

This correspondence is conditional on the realized gate, applies when the aggregate semantic projection is inactive, and does not treat gate-induced interactions as additive concept effects. When projection is active, the mechanism remains exactly reconstructable from the reported contributions, but the aggregate minimum couples semantic concepts and unconditional exact Shapley equivalence is not claimed.

**B.4 Integrated Training and Inference Procedure**

1. **Precompute frozen measurements.** For every development video, uniformly sample frames, compute the generic-safety reading, compute all semantic and deterministic concept measurements, and normalize the concept measurements using statistics from the relevant training partition. The measurement function remains frozen.
2. **Select the operating threshold from out-of-fold predictions.** Partition the development set into five stratified folds. For each fold, fit the risk-assessment parameters for 480 epochs on the other four folds and predict the held-out fold using $p_i^{OOF} = 1 - exp(-\Lambda_i)$. Select the threshold η that maximizes F1 over the combined out-of-fold predictions.

3. **Fit the final development model.** Fit one risk-assessment model on the complete development set for 480 epochs using the unified objective. Lock the learned parameters and the out-of-fold decision threshold $\eta = 0.6123511$ before accessing the test set.

4. **Perform inference and produce faithful outputs.** For each new video, compute the frozen measurements, apply the semantic-budget projection, calculate the generic, mechanism-specific, and total hazards, and obtain $\hat{p}_i = 1 - exp(-\Lambda_i)$. Report the moderation decision, hazard shares, and realized concept contributions from the same forward computation.

The same workflow is restated below in algorithmic form to make explicit how frozen measurement, out-of-fold threshold selection, final fitting, and faithful inference are integrated.

**Algorithm B1. ChildRiskGuard: Joint Training and Inference**

| Line | Procedure |
|---|---|
| | **Input:** development data $D_{dev}$; five folds; fixed measurement function $M$; mechanism blocks $C_r$; activation families $A_{rf}$; contextual attenuation set $E$. |
| | **Output:** fitted risk-assessment parameters θ*, locked operating threshold η, predictions, hazard shares, and concept contributions. |
| 1 | **Precompute frozen raw measurements.** |
| 2 | For each video $V_i$ in $D_{dev}$: |
| 3 | Compute the generic-safety reading $u_i$ and raw concept measurements $C_i$ using $M$. |
| 4 | End for. |
| 5 | Partition $D_{dev}$ into five stratified folds. |
| 6 | For each held-out fold $j$: |
| 7 | Fit $\theta^{-j}$ on the other four folds using the unified objective while keeping M frozen. |
| 8 | Predict the held-out fold with the out-of-fold (OOF) prediction $p_i^{OOF} = 1 - exp(-\Lambda_i)$. |
| 9 | End for. |
| 10 | Select η to maximize F1 over the combined out-of-fold predictions. |
| 11 | Fit θ* on the complete development set and lock η before test evaluation. |
| 12 | For each new video $V_i$: |
| 13 | Compute frozen measurements, apply the final development-set normalization map and measurement statistics, and apply the semantic-budget projection. |
| 14 | Compute generic, mechanism-specific, and total hazards; then compute $p_i$ and the moderation decision. |
| 15 | Output the decision, hazard shares, and realized concept contributions from the same forward computation. |
| 16 | End for. |

Table B1 summarizes the key notation used throughout Appendix B.

**Table B1. Key Model Notation**

| Symbol | Definition |
|---|---|
| $M$ | Fixed measurement function. |
| $u_i$ | Generic-safety reading for video $i$. |
| $c_i$ | Vector of interpretable concept measurements. |
| $s_i$ | Vector of frozen generic-safety evidence channels; used in training only. |
| $H_\theta$ | Trainable risk-assessment function. |
| $\theta$ | Complete set of trainable parameters in $H_\theta$. |
| $R$ | Number of predefined child-specific risk mechanisms. |
| $\lambda_i^G$ | Generic-safety hazard. |
| $\lambda_i^r$ | Hazard of child-specific mechanism $r$. |
| $\lambda^0$ | Input-independent baseline hazard. |
| $\Lambda_i$ | Total hazard. |
| $p_i$ | Child-inappropriateness probability obtained from total hazard. |
| $\eta$ | Moderation decision threshold. |

In the final implementation, $s_i$ contains frozen calibrated visual-sexual, visual-violence, ASR-toxicity, and OCR-toxicity readings, $\tau_{act} = 0.5$, and mechanism terms with fewer than 20 active samples in a minibatch are omitted from that minibatch penalty. Thus, structural identifiability ensures a unique component-level attribution under the conditions of Proposition 2, while $L_{sep}$ softly discourages statistical duplication of generic-safety evidence without imposing strict statistical independence.

**Appendix C Additional Generalizability and Sensitivity Analyses**

**C.1. Generalizability to an Additional Child-Specific Risk Mechanism**

We examine whether the proposed architecture is extensible beyond the initial mechanism set (M1-M3) by introducing an additional, independently grounded child-specific risk mechanism, M4. Following the same procedure used for the primary mechanisms, M4 is defined from external substantive knowledge and operationalized through a separate concept block and mechanism-specific evidence structure. We then add M4 while retaining the existing architecture and M1–M3 specification and re-estimate the risk-assessment stage. Importantly, extending ChildRiskGuard to M4 requires new M4-specific measurements but does not require redesigning the core risk-assessment architecture, modifying the existing M1–M3 specification, or fine-tuning the frozen foundation model. Table C1 reports the substantive grounding of M4 and the resulting change in predictive performance. This analysis tests the generalizability of the mechanism-structuring approach rather than the completeness of any particular risk taxonomy: adding M4 demonstrates whether ChildRiskGuard can accommodate a previously unmodeled risk, but does not imply that four mechanisms exhaust the risks relevant to children.

**Table C1. Generalizability Analysis: Extension to an Additional Child-Specific Risk Mechanism**

| *Panel A. Additional Risk Mechanism* | | | | | |
|---|---|---|---|---|---|
| **Extension** | **Theoretical Basis** | **Child-Specific Risk** | **Design Implication** | **Mechanism** | **Operationalization in ChildRiskGuard** |
| M4: Audiovisual Overstimulation | Children's still-developing attentional filtering and emotion-regulation capacities make them particularly sensitive to intense and rapidly changing audiovisual stimulation. | Rapid editing, frequent auditory onsets, and highly stimulating audiovisual content may induce excessive arousal, attentional overload, and emotional dysregulation. | Add an independent mechanism-specific concept block. | Sensory overstimulation and emotion-regulation risk | M4 combines deterministic measures of shot-cut rate and audio-onset rate with a frozen semantic overstimulation-recall score. A mechanism-specific conjunctive gate activates a nonnegative monotone additive hazard, $\Lambda_{M4}$. |
| *Panel B. Predictive Performance after Mechanism Extension* | | | | | |
| **Model** | | | **F1** | **Precision** | **Recall** |
| ChildRiskGuard + M4 | | | 0.7802 ± 0.0022 | 0.7844 ± 0.0008 | 0.7760 ± 0.0037 |

### C.2. Sensitivity to the Semantic Representation

Table C2 evaluates sensitivity to a key measurement choice in ChildRiskGuard: the semantic representation used to measure the developmental-risk concepts. We replace the semantic readout encoder while keeping the rest of the architecture unchanged. F1 decreases from 0.769 with the reference representation to 0.731 with OpenCLIP and 0.709 with SigLIP, corresponding to drops of 0.038 and 0.060, respectively. These results show that using a semantic representation aligned with the reused safety model provides a measurable performance advantage. At the same time, the performance decline with alternative encoders indicates that the model is sensitive to the source of semantic measurement. Thus, the choice of semantic representation is both an important design component and a boundary condition of the current artifact.

**Table C2. Sensitivity to the Semantic Readout Source**

| Semantic source | F1 | Precision | Recall |
|---|---|---|---|
| CLIP ViT-L/14-336 (Ours) | **0.7689 ± 0.0019** | **0.7940 ± 0.0007** | 0.7453 ± 0.0030 |
| OpenCLIP concept readout | 0.7309 ± 0.0068 | 0.6771 ± 0.0306 | **0.7960 ± 0.0248** |
| SigLIP concept readout | 0.7087 ± 0.0018 | 0.7204 ± 0.0028 | 0.6973 ± 0.0060 |

## Appendix D Theoretical Grounding and Operationalization of AI4CDS Governance

This appendix develops the theoretical basis and operational meaning of the three governance principles introduced in the main text: Graduated Trust, Reversibility, and Auditability. We treat these principles as prescriptive methodological guidance for AI-enabled computational design science (CDS), positioned at the level of nascent prescriptive knowledge. This positioning is consistent with design science research in which design principles and technological rules can constitute nascent prescriptive knowledge whose scope and maturity increase through subsequent application and evaluation (Gregor and Hevner 2013). The derivation below therefore proceeds from established theoretical mechanisms to governance risks created or amplified by constitutive AI participation, then to a prescriptive principle, an observable trigger, and a required researcher response. Evidence from ChildRiskGuard illustrates how the resulting principles operate; it is not, by itself, the theoretical justification for those principles. Accordingly, the case surfaces recurring governance problems and provides instantiation evidence, while established theory explains the underlying mechanisms and disciplines their abstraction into transferable prescriptive principles.

### D.1. Conceptual Basis and Theoretical Derivation

Governance at consequential scientific transitions

AI-enabled CDS requires a distinction between generating a possible research output and authorizing that output as part of the scientific record. We define a scientific commitment as a researcher-authorized decision that promotes a provisional output, alternative, analysis, or interpretation into the research process in a way that constrains subsequent work or supports a scientific claim. Examples include admitting a resource, retaining a consequential design choice, fixing an evaluation specification, accepting a computational result as evidence, or abstracting an observed pattern into a transferable claim. AI may generate candidates for any of these purposes, but generation does not itself constitute scientific commitment.

A consequential transition occurs when an AI-generated or AI-supported output is promoted into, modifies, or causes reconsideration of a scientific commitment in a way that can materially affect downstream decisions or published knowledge claims. Consequence can arise through downstream propagation, evidentiary significance, claim significance, or the cost of correcting an erroneous commitment after subsequent work depends on it. AI4CDS therefore uses consequential transitions as the primary unit of governance, focusing oversight on interactions that affect scientific commitments.

This distinction also separates AI discretion from human scientific authority. AI discretion is the latitude given to AI to select intermediate actions or generate substantive outputs within researcher-defined task boundaries without requiring approval for every intermediate operation. Human scientific authority is the nondelegable responsibility to determine which outputs become scientific commitments and to remain accountable for the validity and scope of the resulting claims. AI4CDS therefore requires human authority at scientific commitment points while allowing AI discretion over intermediate computational steps.

**Graduated Trust**

Research on trust in automation distinguishes productive reliance from both overreliance and underreliance. Lee and See (2004) argue that trust is consequential because it guides reliance when automation is complex and cannot be completely understood, whereas Fügener et al. (2022) show that productive human–AI delegation is difficult when humans lack the metaknowledge needed to assess relative capabilities. In scientific settings, these problems are compounded by the possibility that fluent AI outputs create an appearance of understanding that exceeds the researcher's actual epistemic basis (Messeri and Crockett 2024).

AI4CDS translates this problem into Graduated Trust: AI discretion should be calibrated to the explicitness of task constraints, the feasibility of independent verification, and the scientific consequence of error. More explicit constraints and stronger independent verification permit greater AI discretion; open-ended judgment, weak verification, or high downstream propagation require stronger researcher control. Although the principle is named Graduated Trust, its operational target is observable reliance and delegation, expressed through task-level decisions about AI discretion.

Graduated Trust is therefore task-contingent, not phase-monotonic. A tightly specified implementation or computation may permit substantial AI discretion, whereas an open-ended interpretation or abstraction may require strong human authority even when it occurs later in the research process. Phase location can affect task characteristics, but phase number itself does not determine appropriate delegation. Regardless of AI discretion, researchers retain authority over consequential scientific commitments.

**Reversibility**

Design fixation describes the tendency for exposure to an initial solution to constrain subsequent design search and encourage continued adherence to a limited set of ideas (Jansson and Smith 1991). This risk is particularly relevant when AI can rapidly produce plausible and seemingly complete designs that are then embedded in resources, code, evaluations, and manuscript claims. Design science also treats construction and evaluation as iterative: evaluation provides feedback to construction, and assumptions or design choices may need to be revised when subsequent evidence exposes deficiencies or changed conditions (Hevner et al. 2004).

AI4CDS responds through Reversibility: an earlier scientific commitment must remain reopenable when credible later evidence materially challenges the assumptions, requirements, measurements, or evidence on which that commitment depended. The trigger is not simply the availability of a new alternative but evidence that calls the earlier justification into question. When this occurs, researchers return to the earliest materially implicated commitment and explicitly decide whether to retain, revise, replace, or reject it.

Reversibility does not prohibit downstream modification. Rather, a downstream modification must not substitute for reconsidering an upstream commitment whose justification has been materially contradicted. After that reconsideration, researchers may determine that the original commitment remains justified and that a downstream modification is appropriate. Reversibility is also bidirectional: it protects against retaining a poorly supported choice, but it also protects against prematurely abandoning a choice because of weak, unstable, or isolated contrary evidence.

**Auditability**

Computerized decision making can diffuse responsibility and make it difficult to identify how consequential outcomes arose and who remains accountable for them (Nissenbaum 1996). These concerns become especially salient when AI participates in research because authorship, transformation, and decision boundaries may become difficult to reconstruct. Recent guidance for AI-enabled IS research accordingly emphasizes provenance and verification alongside disclosure, including records of AI contributions, workflow histories, model and data states, and human decision gates at consequential points (Gopal et al. 2025).

AI4CDS translates these concerns into Auditability: consequential transitions should leave an inspectable record sufficient for an informed third party to determine what AI contributed, what researchers decided, what evidence or criteria supported the decision, and how that decision affected the resulting artifact or claim. At minimum, the record should identify the relevant context or input, the substantive AI contribution, the human disposition, the decision basis, and the scientific consequence. Model or tool versions, prompts, execution specifications, data states, and verification procedures should additionally be preserved when they are material to reconstructing the transition.

Auditability therefore scales documentation to the scientific consequence of an AI-supported interaction. Routine formatting, unused suggestions, and other assistance that does not alter a scientific commitment need not receive the same documentation as a retained design decision, reported result, or transferable claim. Auditability also does not require access to a model's hidden internal reasoning. The auditable scientific object is the observable research process: the task and inputs supplied to AI, its externally available contribution, the researcher's decision, the supporting evidence, and the downstream consequence.

### D.2. Operationalization at Consequential Transitions

The three principles govern different aspects of the same transition. Graduated Trust determines how much discretion AI may receive before the transition; Reversibility determines what happens if later evidence undermines the resulting commitment; and Auditability determines what record must remain so that the transition and any later reconsideration can be evaluated. Table D1 translates each principle into an observable assessment and required response.

**Table D1. Theory-Grounded Operationalization of AI4CDS Governance**

| Principle | Theoretical mechanism and governance risk | Operational trigger or assessment | Required response | ChildRiskGuard illustration |
|---|---|---|---|---|
| Graduated Trust | Appropriate reliance and delegation depend on the ability to assess task and agent capabilities (Lee and See 2004; Fügener et al. 2022). AI creates risk of both premature reliance and failure to exploit useful capabilities. | Assess task-constraint explicitness, independent verification feasibility, and the propagation or scientific consequence of error. | Increase AI discretion when constraints and verification are strong and consequences are bounded; reduce discretion as judgment becomes more open-ended, verification weaker, or consequences greater. Researchers authorize all scientific commitments. | AI broadly generated literature and design alternatives, while researchers determined scope, admissibility, and design commitments. AI received greater implementation and diagnostic discretion after specifications were fixed, but researchers verified outputs before accepting results or interpretations. |
| Reversibility | Initial solutions can induce design fixation (Jansson and Smith 1991), while iterative design requires evaluation evidence to feed back into construction (Hevner et al. 2004). Rapid AI-supported development can make provisional choices increasingly embedded in later work. | Credible later evidence materially challenges the assumptions, requirements, measurements, or evidentiary basis of an earlier commitment. | Reopen the earliest materially implicated commitment and explicitly retain, revise, replace, or reject it. A downstream modification cannot substitute for reassessing a contradicted upstream commitment. | Candidate mechanisms were reduced when measurement and development evidence did not sustain their differentiation; a dangerous-objects mechanism was rejected for insufficient measurement specificity; gate structures were revised. Conversely, a recommendation to remove a mechanism based on one split was not adopted when broader evidence did not support removal. |
| Auditability | Computerization can diffuse accountability (Nissenbaum 1996), while responsible AI-enabled research requires provenance and verification records (Gopal et al. 2025). | An AI-supported output becomes, modifies, or causes reconsideration of a consequential scientific commitment. | Preserve a consequence-proportional record of the input or context, AI contribution, human disposition, decision basis, and downstream scientific consequence, together with technical metadata needed for verification where relevant. | Appendix A documents AI and researcher roles across the five phases. Table A2 records consequential design candidates, their origins, dispositions, and bases; Table A3 records evaluation issues and researcher corrections, rejections, or qualifications. Appendix A.5 links consequential decisions to evidence when candidate transferable claims are assessed. |

The ChildRiskGuard record also illustrates why these principles should be applied at transitions rather than assigned permanently to phases. Table A2 contains AI-originated proposals that were accepted, modified, or rejected after theoretical, measurement, or development-data review. Table A3 similarly records cases in which implementation or interpretation was corrected, qualified, or rejected before it became part of the scientific account. Appendix A.5 then uses the accumulated cross-phase record when candidate design and governance knowledge is abstracted. Thus, increasing AI participation in an activity does not transfer scientific authority to AI: the relevant governance question remains how a provisional contribution crosses into a consequential commitment.

### D.3. Differentiated Reproducibility as an Auditability Standard

Auditability also requires that the form of verification match the type of research activity being audited. AI-enabled CDS combines deterministic computation, stochastic model fitting, generative search and diagnosis, and formal analytical claims. Requiring the same form of reproduction from each activity would either impose an infeasible exact-reproduction requirement on stochastic AI-assisted work or accept insufficiently precise evidence for computational procedures that can be directly rerun. We therefore use differentiated reproducibility to match each component to the form of independent verification appropriate to its computational determinism, stochasticity, process dependence, and claim type.

The four categories below are distinct verification standards. Each represents a different form of rigor appropriate to a different research activity.

**Table D2. Differentiated Verification Standards for AI-Enabled CDS**

| Verification standard | Appropriate research activity | Verification requirement | ChildRiskGuard instantiation |
|---|---|---|---|
| Computational Reproducibility | Fixed or effectively deterministic computational procedures | A qualified researcher can re-execute the specified procedure on the same inputs under an equivalent computational environment and obtain numerically equivalent outputs within an appropriate tolerance. Code, data, configurations, relevant software versions, and fixed parameters should be preserved. | Appendix B specifies the frozen measurement procedures, deterministic extractors, normalization rules, architecture, inference computation, and integrated training/inference workflow. Deterministic and frozen components can therefore be rerun from their stated inputs and specifications. |
| Statistical Reproducibility | Stochastic computation, including model fitting and procedures affected by random initialization or sampling | Repeated executions under the specified procedure produce results consistent with the reported variation. The record should preserve the number of runs, randomization procedure or seeds where applicable, and summary statistics needed to assess stability. | Learned models were evaluated across five random seeds and results are reported with means and standard deviations. The very small variation observed for ChildRiskGuard was checked rather than assumed to be a reporting error. |
| Procedural Reconstructability | Stochastic and context-dependent AI-assisted activities such as literature synthesis, candidate generation, design critique, diagnosis, and abstraction | A qualified researcher can reconstruct what task was attempted, the inputs and constraints supplied, the relevant alternatives considered, the consequential decisions made, and the verification logic through which AI outputs did or did not become scientific commitments. Exact regeneration of the same AI text, candidate sequence, or final artifact is not required. | Appendix A.1–A.5 documents task constraints, AI and researcher roles, verification practices, design alternatives, corrections, and abstraction procedures. Tables A2 and A3 make consequential dispositions and corrections inspectable without requiring regeneration of the original stochastic AI interactions. |
| Claim Verifiability | Formal propositions, analytical properties, or other claims whose validity follows from stated assumptions and derivations | A qualified researcher can independently assess whether the claim follows from the stated assumptions, definitions, artifact structure, and derivation without reproducing the AI interaction that may have supported its development. | Appendix B.3 states the relevant assumptions and provides the derivations and proofs for Propositions 1–4 and the associated formal faithfulness properties, permitting direct independent scrutiny of the claims. |

Differentiated reproducibility changes the object of verification without relaxing the requirement for rigor. For deterministic computation, the object is computational execution; for stochastic computation, it is the stability of results under repeated execution; for AI-assisted search and judgment, it is the reconstructability of the scientific decision process; and for formal claims, it is the validity of the derivation. In particular, procedural reconstructability does not require another researcher to obtain the same generative output. It requires enough provenance to evaluate how AI-supported alternatives were transformed into researcher-authorized scientific commitments.

### D.4. Inter-Principle Relationships, Tensions, and Boundary Conditions

The three principles are complementary but not costless. First, Graduated Trust and Reversibility jointly prevent accumulated reliance from becoming irreversible commitment. Successful prior AI performance may justify greater discretion on comparable, well-specified tasks, but it does not immunize an earlier decision from later contradictory evidence. When credible evidence challenges a scientific commitment, the Reversibility trigger takes precedence over accumulated confidence in either the AI or the existing design.

Second, Reversibility depends on Auditability. Reconsidering an earlier commitment requires knowing why it was made, what evidence supported it, and what subsequent decisions depend on it. Reopening a consequential commitment therefore creates a corresponding audit obligation: the record should identify the new evidence and whether the commitment was retained, revised, replaced, or rejected. This requirement does not imply documenting every minor iteration; full documentation applies to consequential reversals.

Third, Auditability creates a practical tension between completeness and feasibility. Recording every prompt, intermediate suggestion, formatting change, or routine coding interaction can generate large records without proportionate scientific value. AI4CDS addresses this tension through proportionality: governance effort should increase with the scientific consequence, verification difficulty, and propagation potential of an AI-supported decision. Routine assistance can therefore receive lightweight documentation, whereas transitions that determine admitted evidence, artifact structure, reported results, or transferable claims require a substantially stronger record.

Together, these relationships establish a priority rule for AI-enabled CDS: at consequential scientific transitions, validity and accountability take precedence over efficiency or convenience. Greater AI discretion is appropriate where researchers can constrain and independently verify the task; prior commitments remain reopenable when later evidence challenges their justification; and consequential transitions remain traceable to observable AI contributions, researcher decisions, and supporting evidence.

These principles define prescriptive governance for AI4CDS within explicit boundary conditions. Their application depends on the consequence of the task, the availability of independent verification, the properties of the AI system, and the research setting. ChildRiskGuard demonstrates that the three principles can be instantiated within a consequential computational design project and connected to an inspectable process record. Further studies across different CDS problems, AI capabilities, research teams, and epistemic settings are needed to test, refine, and delimit the conditions under which these principles provide effective governance.